\documentclass[manuscript]{acmart}
\usepackage{acro}[=v2]
\setcopyright{acmlicensed}
\copyrightyear{0}
\acmYear{0}
\acmDOI{XXXXXXX.XXXXXXX}
\acmJournal{CSUR}
\acmVolume{0}
\acmNumber{0}
\acmArticle{0}
\acmMonth{0}
\usepackage{multirow}
\usepackage{graphicx}
\usepackage{xcolor}
\usepackage{multicol}
\usepackage{amsmath}
\usepackage{multirow}

\usepackage[font={footnotesize}]{subcaption}
\usepackage{cancel}
\usepackage[flushmargin]{footmisc}
\let\oldnl\nl
\newcommand{\nonl}{\renewcommand{\nl}{\let\nl\oldnl}}

\usepackage{graphicx}
\usepackage[normalem]{ulem}
\usepackage{ragged2e}
\usepackage{hyperref}
\usepackage[table]{xcolor}
\usepackage{threeparttable}
\usepackage{adjustbox}
\usepackage{soul}
\usepackage{enumitem}
\renewcommand\footnotetextcopyrightpermission[1]{} 
\usepackage{microtype}
\setlist{leftmargin=*, topsep=0pt, partopsep=0pt, parsep=0pt, labelsep=0.7ex}
\begin{document}
\title{Upholding Robustness in Federated Learning: Trends,  Emerging Strategies, and Research Opportunities}
\author{Pravija Raj P V}
\email{p20230903@dubai.bits-pilani.ac.in}
\orcid{0000-0002-9286-701X}
\affiliation{
  \institution{Department of Computer Science and Engineering, BITS Pilani Dubai Campus}
  \city{Dubai}
  \country{UAE}
}
\author{Ashish Gupta}
\email{ashish@dubai.bits-pilani.ac.in}
\affiliation{
   \institution{Department of Computer Science and Engineering, BITS Pilani Dubai Campus}
  \city{Dubai}
  \country{UAE}
}
\author{Andrea Augello}
\email{andrea.augello01@unipa.it}
\affiliation{
  \institution{Department of Engineering, University of Palermo}
  \city{Palermo}
  \country{Italy}
}
\author{Sajal K. Das}
\email{sajal@mst.edu}
\affiliation{
   \institution{Department of Computer Science, Missouri University of Science and Technology}
  \city{Rolla}
  \country{USA}
}
\renewcommand{\shortauthors}{Pravija et al.}
\begin{abstract}
While Federated Learning (FL) has been widely adopted for protecting user privacy in machine learning, it remains vulnerable to various robustness challenges, including performance-impairment risks, information-stealing threats, and aggregation vulnerabilities. This work offers a holistic synthesis of FL robustness along three tightly coupled angles: (i) a threat-centric view of robustness that categorizes the multifaceted attack surfaces, (ii) a structured taxonomy of robust aggregation strategies distinguishing outcome-centric approaches from security-centric strategies, and (iii) a layered taxonomy of defensive strategies. We rigorously examine current evaluation practices for FL robustness and identify major applications and open research challenges to guide future research.
\end{abstract}
\maketitle
\section{Introduction} \label{intro}
Federated Learning (FL) is causing a paradigm shift in the field of Artificial Intelligence (AI) by enabling decentralized model training across numerous devices while keeping data local~\cite{wang2024invariant, zhang2026secure, gupta2023performance}.
The proliferation of intelligent devices and applications that generate vast, decentralized, and heterogeneous data presents tremendous opportunities for scientific and technological innovation.
FL offers substantial advantages over traditional centralized architectures by facilitating real-time analytics and collaborative knowledge aggregation without centralizing sensitive information~\cite{javeed2024quantum, hamouda2023ppss, jiang2020poisoning, zhu2021distributed}.
Consequently, the framework is rapidly gaining traction across diverse sectors, including finance, healthcare, Internet of Things (IoT), and smart cities~\cite{mothukuri2021survey}. 
The decentralized nature of FL makes it prone to threats that compromise collaborative training~\cite{hao2023robust, 9878489, 9798077}. 
Real-world deployments face adversarial and heterogeneous environments, characterized by diverse participant devices with varying computational and communication resources, and Non-Independent and Identically Distributed (Non-IID) and imbalanced data distributions~\cite{de2026federated}. Further, either some clients or the server may act covertly or maliciously, and communication channels may leak sensitive information. Under these circumstances, ensuring FL robustness involves addressing potential threats; therefore, it is vital to continuously assess the model for vulnerabilities~\cite{Wan_2023, jeong2024fedcc}.
Tackling these challenges necessitates multi-faceted solutions that extend beyond discrete defenses.
In practice, robust FL demands careful and effective handling of multiple aspects, including threats, aggregation robustness, and defensive measures. 
Furthermore, the inconsistency in evaluation trends for robust FL with research using different datasets, partitioning choices, attack settings, mitigation strategies, and metrics makes it complicated to assess methods or verify their particular relevance in real-life circumstances.
In this paper, {\em we review and systematically analyze the challenges facing FL and assess the current solutions from diverse perspectives, including threats, aggregation strategies, and defenses, which a researcher should contemplate for upholding robustness.} To facilitate an easy understanding, we begin by outlining the crucial factors for robustness in FL.
\noindent $\bullet$ \textbf{Crucial Factors for Robustness:} Unlike conventional distributed optimization, federated models demand a heavy focus on reliability and robustness. 
Key factors that have a substantial impact on the FL robustness can be assessed from multiple angles.
\noindent $\bullet$  \textit{Adversarial threats:}  Malicious participants may manipulate data, alter model updates, or initiate backdoor attacks during training to undermine the global model~\cite{xie2020dba}. Detection and mitigation of adversarial threats at the server side can greatly reduce malicious impact~\cite{gong2022coordinated, Watermarking2024}. Resilience is the key to strong FL frameworks; they should be capable of resisting malevolent attacks~\cite {gupta2022long} and greedy participants~\cite{augello2024tackling}.
\noindent $\bullet$  \textit{Aggregation process:} Traditional FL aggregation solutions, such as FedAvg~\cite{mcmahan2017communication}, are vulnerable to deceitful updates, uneven data, and outliers~\cite{wang2024privacy, guo2023fast, tang2024flexible}. Furthermore, a fragile aggregation approach on the server, if faced by adversaries, could detrimentally affect the whole training process. In contrast, adaptive and customized approaches can achieve robust models~\cite{li2019rsa, tan2022fedproto, wu2023hiflash}.
\noindent $\bullet$  \textit{Defensive mechanisms:} Detecting the presence of adversaries and excluding them from the aggregation process is vital to developing a reliable FL model; consequently, different defensive strategies~\cite{gupta2022long,wu2022threats, yu2023untargeted} have been devised, which demand consolidated discussion and comparative analysis to steer future research toward robust FL.  
\subsection{Motivation} \label{rr}
The widespread adoption of FL has resulted in many survey papers to handle inherent challenges and diverse threats, and devise appropriate mitigative measures and defenses~\cite{li2024contribution, de2026federated, tang2024flexible}. 
This section inspects these efforts through a critical lens, discussing their major insights and gaps, underscoring the need for our work in this area.
To begin with,~\cite{lianga2023survey} provides an analysis of client-side threats, with a key focus on how model and data poisoning can disrupt the system. Meanwhile, a thorough discussion of attack vectors, such as evasion, model inversion, and backdoor tactics, is provided in~\cite{sikandar2023detailed}.
These studies also critically assess the efficacy of countermeasures such as adversarial training and Differential Privacy (DP) against these serious threats.
While the studies~\cite{mothukuri2021survey, han2024privacy} examine security, privacy, and deployment risks, the authors in~\cite{lyu2020threats} present a taxonomy that primarily includes threat models and two main attack types (poisoning and inference). In addition to a taxonomy of attacks, the work in~\cite{liu2022threats} analyzed the potential defenses and summarized their shortcomings.  
A survey analyzing the FL from the differential privacy aspect is presented in~\cite{fu2024differentially}
In~\cite{zhang2023survey}, the authors noted that an adversary's access levels are crucial for fully understanding the impact of rule exposure.  The work~\cite{wu2022threats} reviewed different attack types based on the potential roles attackers might play in FL.  From a security perspective, a categorization is proposed in~\cite{chen2022federated} based on the security properties: integrity, confidentiality, and availability.  A categorization framework that differentiates between vertical, horizontal, and transfer learning-based FL is proposed in~\cite{yang2019federated}. 
The study in~\cite{lim2020federated} focused on enhancing deployments in distributed computing frameworks
by highlighting the challenges of FL implementation in mobile edge computing. A detailed review of state-of-the-art attacks that compromise privacy using membership attacks and utility, and their defenses, is given in~\cite{li2024threats, bai2024membership}.  In addition, the systematic review~\cite{uddin2025systematic} provides a categorical analysis of state-of-the-art FL defenses, based on their underlying principles and techniques.\\
\noindent $\bullet$ \textbf{Major gaps in the prior surveys:} After a thorough study, we discovered the following gaps: Despite rapid progress, past studies generally inspect robustness from a single dominant angle, such as security,  non-IID data, privacy, or aggregation challenges, resulting in a {\em fragmented insight on robustness} in practical FL systems. Most existing reviews primarily focus on specific adversary types, {\em inspecting each category in isolation}, which restricts their ability to deliver a holistic view of robustness across the crucial factors of FL. While prior works have devised valuable taxonomies, they didn't go much beyond conventional security threats and thus often {\em fall short of delivering a detailed analysis} over diverse robustness perspectives.  
 To the best of our knowledge,  most existing surveys do not adequately orchestrate and assess the multifaceted constituents of robustness over the whole FL pipeline under a unified perspective. This highlights the pressing need to unite these research strands and systematically examine FL through the {\em lens of robustness}, enabling a holistic review that can deliver a coherent reference for upholding FL robustness.
\subsection{Survey Scope}
While addressing the above-mentioned gaps, this paper offers a deeper, more insightful, and end-to-end understanding of FL robustness along three tightly coupled angles.
The major contributions are:
\begin{itemize}
    \item This study presents a unified, multi-layer, threat-centric view of FL robustness by categorizing performance-impairment and information-stealing attacks across different layers of the FL pipeline. We expose and discuss the multifaceted attack surfaces affecting FL robustness that previous studies handled separately. 
    \item We propose an innovative multi-level taxonomy of robust aggregation strategies that distinguishes outcome-centric from security-centric approaches, emphasizes their robustness assumptions, demonstrates their interactions with defensive mechanisms, and offers a holistic design space for robust aggregation.
    \item This work introduces a layered taxonomy of defensive strategies categorized by their stage of action (e.g., training data, local updates, global model, and server/coordination). 
 \item Finally, we review current evaluation trends, covering commonly used data-partitioning schemes, benchmark datasets, and attack configurations; identify robustness gaps in the existing literature; outline key application scenarios; and highlight open research challenges to guide future research on developing robust FL systems.
\end{itemize}
\begin{figure}[!htb]
    \centering
   \includegraphics[width=0.75\linewidth]{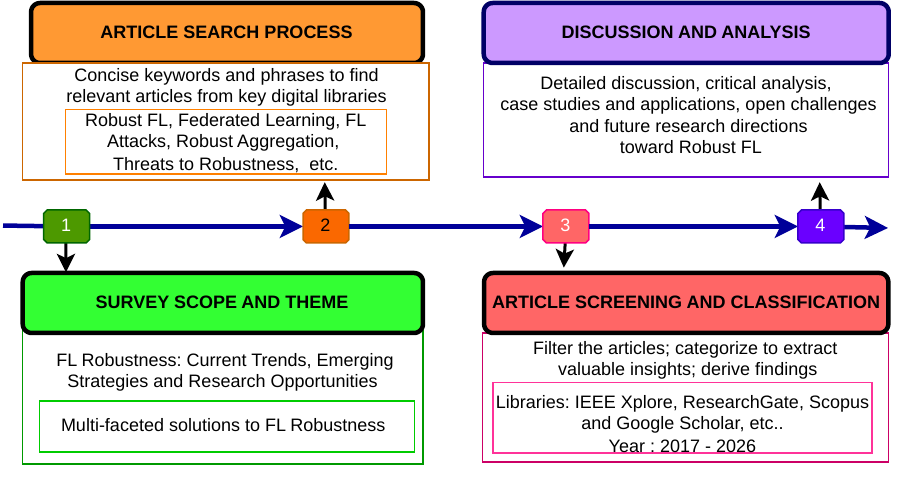}  \vspace{-10pt}
    \caption{Research protocol steps.}
   \label{fig:scheme}
   \Description{The structured approach for the survey.}
   \vspace{-0.2in}
\end{figure}
\subsection{Research Protocol}
To ensure comprehensive coverage of the literature on robust FL, we follow a structured approach as shown in Fig.~\ref{fig:scheme}.
\noindent $\bullet$ {\textit{Articles search process:}}
 We start by filtering papers from key digital libraries such as IEEE Xplore, ResearchGate, Scopus, and Google Scholar. The search and selection process employed research-specific keywords such as:  \textit{Federated Learning}, \textit{Robust FL}, \textit{FL Robustness},  \textit{Robust Aggregation}, \textit{Threats to Robustness}, \textit{FL Attacks}, and \textit{FL Security}. 
Search queries are constructed using carefully selected keywords and brief phrases, along with Boolean operators, such as AND and OR. 
A detailed search was performed across the chosen databases, focusing on titles, abstracts, and keywords spanning from 2017 to 2026. The primary studies are selected based on quality and after screening their abstracts or full-length papers.  Duplicates, irrelevant research, and poor-quality papers are eliminated.   The selected papers are systematically categorized, enabling a more focused review to identify key insights and trends.
\noindent $\bullet$ {\textit{Selection and inclusion criteria:}} To guarantee relevance, quality, and impact, the research papers are carefully screened using a structured evaluation checklist. Major selection and inclusion criteria includes: (i) high-impact journals or conferences with significant citations indicating influence; (ii) concise problem formulations with precise explanations; (iii) understandable and well-articulated presentation and discussions; (iv) technical depth in terms of detailed algorithms and implementations; (v) novel solutions addressing FL threats while maintaining viable robustness guarantees; (vi) detailed comparative discussion with state-of-the-art; and (vii) evidence-based conclusions, to ensure the inclusion of the most relevant literature. 
\noindent {\bf Survey organization:} Section~\ref{sec:threats} dives into the diverse threats to FL robustness. We start with an overview of robustness challenges and then classify attacks. Section~\ref{sec:aggr} introduces a taxonomy of robust aggregation strategies, categorizing and analyzing research approaches to ensure model integrity and robust performance. Section~\ref{sec:def} details defensive strategies for FL robustness using an insightful taxonomy. 
Section~\ref{sec:cri} delivers the experimental evaluation trends. Section~\ref{sec:future} offers a summary discussion, sharing insights into emerging trends, major application scenarios, and outlining open challenges and future research directions. 
\section{Threats to Robustness in FL} \label{sec:threats}
This section critically reviews the threats to FL robustness identified in existing studies. In FL, robustness refers to the system's ability to either \textbf{remain unaffected by threats and adversarial conditions} or \textbf{recover automatically} from them while maintaining reliable performance. 
Prior works~\cite{gupta2022long, li2024threats, yu2023untargeted, bagdasaryan2020backdoor, jere2020taxonomy, sun2021data, chen2022federated, arevalo2024task, wei2023covert, zhou2021deep, wang2020attack, lyu2023poisoning, jeong2024fedcc, zhang2023denial, sun2023attacking, tolpegin2020data, jiang2020poisoning} highlighted FL’s vulnerability to a variety of underexplored threats.
After presenting the essentials of FL training, we divide potential threats into two insightful categories: (i) {\bf performance-impairment threats} -- that cause model performance to drop (either due to a malicious actor or structural conditions), affecting the FL robustness, and (ii) {\bf information-stealing threats} -- that aim to extract sensitive information, posing risks to privacy and security.      
\subsection{FL Training: The Essentials}
\begin{figure}[!htb]
    \centering
    \includegraphics[width=\linewidth]{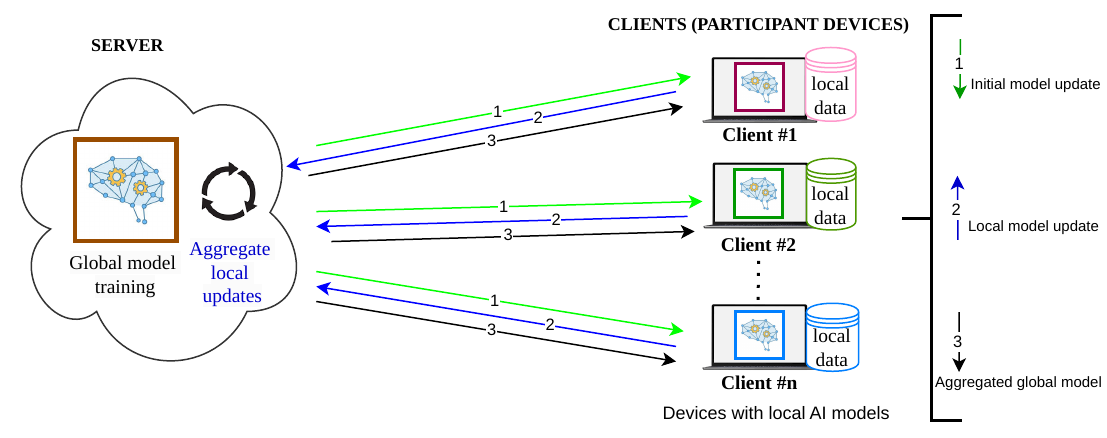} 
    \vspace{-0.3in}
    \caption{Typical FL operational process.}
    \Description{Working of a general FL system.}
   \label{fig:flprocess}
\end{figure}

Fig.~\ref{fig:flprocess} portrays a typical FL setup.
Collaborative model training across multiple participants enables the creation of a global model while protecting data privacy. FL involves two main entities: participants (often called clients) who train ML models using their personal data, and an aggregator server that combines the local models into a global model. The key phases include: {\em training phase}, where clients collaboratively train the model by exchanging local updates without disclosing their data, and {\em inference phase}, where clients apply the trained global model to new data samples. 
The training process is decentralized, implying that the central server (aggregator) has no access to or control over clients' data. Let $P = \{p_1, p_2, \ldots, p_n\} $ represent the set of $n$ participants, who download parameters (model) from the server. Each participant $p_i$ utilizes its private local dataset $D_i$ to train a local model $\zeta^r_i$ during each communication round $r$. 
In particular, each client performs $\tau$ local epochs, and then the updated parameters are uploaded to the server for aggregation to refine the global model $\zeta^r_G$. Following the aggregation, the global model $\zeta_G^r$ is sent to all participants. 
Let $n$ denote the total number of participants, and the weight coefficients satisfy $\sum_{p=1}^{n} \varrho_p = 1$ with $\varrho_p \geq 0$. The local objective function of participant $p$ is defined as: $
Fl_p(\varphi) = \frac{1}{n_p} \sum_{q=1}^{n_p} fl_q(\varphi; x_q, y_q)$,  
where $n_p$ represents the number of local data samples, and $fl_q(\varphi; x_q, y_q)$ is the corresponding loss function.  
And, the server aims to: 
\begin{math}
    \min_\varphi F_G(\varphi),  \text{where } F_G(\varphi) := \sum_{p=1}^{n} \varrho_p Fl_p(\varphi).
\end{math}
\begin{table}[!htb]
\centering
\small
\caption{Adversarial roles and involvements in FL.} 
\vspace{-0.1in}
\label{table:role}
\Description{Involvement of adversaries and their roles in FL}
\renewcommand*{\arraystretch}{1}
\resizebox{0.9\linewidth}{!}
{
\begin{tabular}{|m{0.5cm}|m{15cm}|}
\hline
\rowcolor[HTML]{8c8c8c} \textbf{Role} & \textbf{Involvement in FL} \\
\hline
{\rotatebox{90}{\textbf{Participant}}} &  \begin{itemize}
                \item  Corrupt or replace local model updates.
              \item  Observe the global model.
            \item  Control local training
            (optimize loss function and hyperparameters (learning rate, local epochs, batch size)).
              \item  Coordinate attacks to corrupt or influence the global model.
            \end{itemize} \\ \hline
{\rotatebox{90}{\textbf{Server}}} &  \begin{itemize}
                \item Directly inspect or manipulate global model parameters  
                (but has no access to train or test data).
                \item Inspect local model updates of participants in the absence of secure aggregation mechanisms. 
                \item Inject an attack over the aggregation mechanism.
            \end{itemize}
    \\ \hline
{\rotatebox{90}{\textbf{External}}}     &   \begin{itemize}
                \item Intercept communication between entities.
                \item Perform inference attacks.
                \item Create a malicious model replica.
                \item End users of the deployed service can act as adversaries (they have access to the final trained model).
            \end{itemize}  \\ \hline
\end{tabular}
}
\vspace{-0.25in}
\end{table}
\subsection{Security Challenges in FL}
\begin{figure}[!htb]
    \centering
    \vspace{-0.1in}
\includegraphics[width=\linewidth]{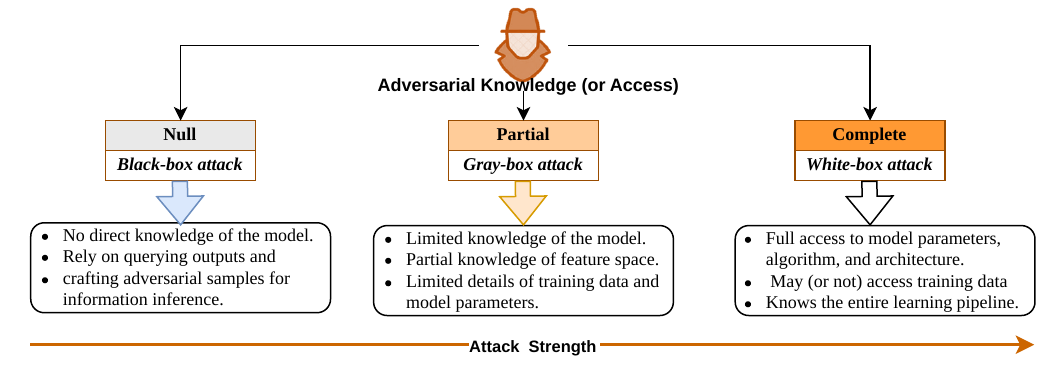}
\vspace{-0.3in}
    \caption{Attack strength based on adversarial knowledge levels.
    } 
    \vspace{0.1in}
    \Description{Shows how attack strength varies based on adversarial knowledge levels}
   \label{fig:know}
\end{figure}

Adversaries can infiltrate the FL framework and pose serious security challenges by exploiting vulnerabilities, such as manipulating training parameters, altering the aggregated model, and distorting learning outcomes.
While localized models and raw training data represent major vulnerabilities on the client side, the integrity of aggregated parameters or gradients can become targets on the server side, particularly in the deployment/inference stages~\cite{yu2023untargeted, bagdasaryan2020backdoor, jere2020taxonomy, sun2021data, jiang2020poisoning}.
An adversary can get involved in the FL process by either compromising a participant or the server, or even acting externally, as discussed in Table~\ref{table:role}. The adversary's capabilities depend on their role in the learning process.
Adversaries can have different levels of knowledge (\textbf{null, partial, complete}) and access to the model based on their role.
This knowledge can include training data, feature space, learning methods, parameters, and even the cost function. Fig.\ref{fig:know} illustrates the levels of access adversaries might possess. 
Through an iterative learning process, an adversary can evolve from a black-box (limited knowledge) to a white-box (complete knowledge) scenario. 
Further, based on the adversary's role and its target, we summarize the general classes of attacks in Table~\ref{tab:fl_attacks}. 
Attacks can be broadly categorized into four types based on their objectives and targets: 
\textbf{(i) Targeted attack:} It aims to degrade the model's performance on a specific task while leaving other tasks unaffected~\cite{mothukuri2021survey}. 
\textbf{(ii) Untargeted attack:} It reduces the model's overall performance or disrupts the model's convergence.
\textbf{(iii) Inference attack:} It aims to extract sensitive information from the model, such as properties of the training data, membership of specific data points, or class information.
\textbf{(iv) Reconstruction attack:} It attempts to recover the actual training samples by analyzing the model's parameters or outputs, causing a serious threat to data privacy.
\begin{table}[!htb]
\centering
\tiny
\renewcommand*{\arraystretch}{1}
\caption{Overview of Adversarial Attacks. }
\label{tab:fl_attacks}
\resizebox{0.8\linewidth}{!}
{
\begin{threeparttable}
\begin{tabular}{|l|c|c|c|c|c|c|c|}
\hline
\rowcolor[HTML]{8c8c8c} \textbf{Class} & \multicolumn{2}{c|}{\textbf{Adversarial Role}} & \multicolumn{2}{c|}{\textbf{Specific Target}} & \multicolumn{2}{c|}{\textbf{Execution Rounds}} & \textbf{AK} \\ 
\cline{2-7}
\rowcolor[HTML]{8c8c8c} & Participant & Server & Data & Model  & One & Many & \\ 
\hline
Targeted &  $\checkmark$ & $\times $& $\times$ & $\checkmark$ & $\checkmark$ & $\checkmark$ & $\checkmark$  \\ 
\hline
Untargeted &  $\checkmark$ & $\times$& $\times$ & $\checkmark$ & $\checkmark$ & $\checkmark$ & $\checkmark$ \\ 
\hline
Inference (P)  & $\checkmark$ & $\checkmark$ & $ \checkmark$ & $\times$ &  $\times$ & $\checkmark$ &$ \checkmark $\\ 
\hline
Inference (M)  & $\checkmark$ & $\checkmark$ &$ \checkmark$ & $\times$ &  $\times$ & $\checkmark$ &$ \checkmark $\\ 
\hline
Inference (C)  & $\checkmark$ & $\checkmark$ & $ \checkmark$ & $\times$ &  $\times$ & $\checkmark$ &$ \checkmark $\\ 
\hline
Reconstruction &$\checkmark$ & $\checkmark$ &  $\checkmark$ & $\times$ &  $\checkmark$ & $\checkmark$ & $\times$ \\ 
\hline
\end{tabular}
\begin{tablenotes}
\item AK (Additional Knowledge), (P) Properties,  (M) Membership, and (C) Class
\end{tablenotes}
\end{threeparttable}
}
\vspace{-0.2in}
\end{table}
\subsection{Performance-Impairment Threats}
\begin{figure}[!htb]
    \centering
    \vspace{-0.1in}
   \includegraphics[width=0.9\linewidth]{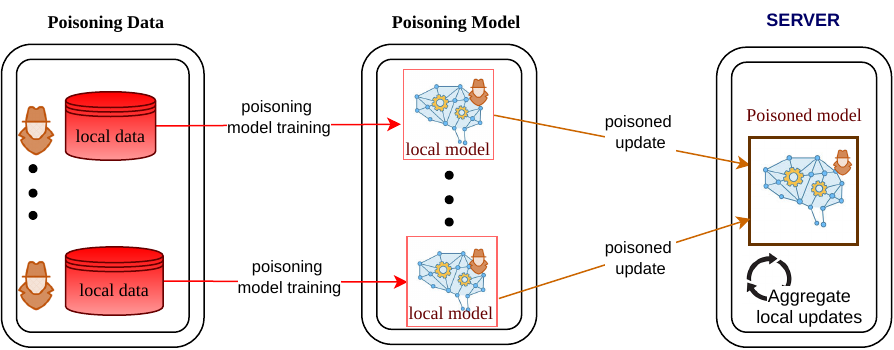}
    \caption{Poisoning-induced attacks in FL. An adversary can poison either the training data or the local model.}
    \Description{Example of attacks induced by poisoning data or model in FL}
   \label{fig:poi}
\end{figure}

The threats to disrupt global model performance, either directly or indirectly, mostly occur during the training phase and are categorized based on adversarial objectives~\cite{wei2023covert, zhou2021deep, wang2020attack, gupta2023performance}.  
They are often implemented by poisoning the data or model~\cite{lyu2023poisoning, jeong2024fedcc, zhang2023denial}, as shown in Fig.~\ref{fig:poi}.
 Data poisoning involves training or inference data manipulation to indirectly compromise a model, whereas model poisoning directly alters the learning process to corrupt it. Additionally, free-riders, attacks on the communication channel, and denial-of-service attacks significantly impact FL performance.   
\subsubsection{Poisoning Attacks}
\begin{figure}[!htb]
    \centering
    \vspace{-0.1in}
   \includegraphics[width=0.9\linewidth]{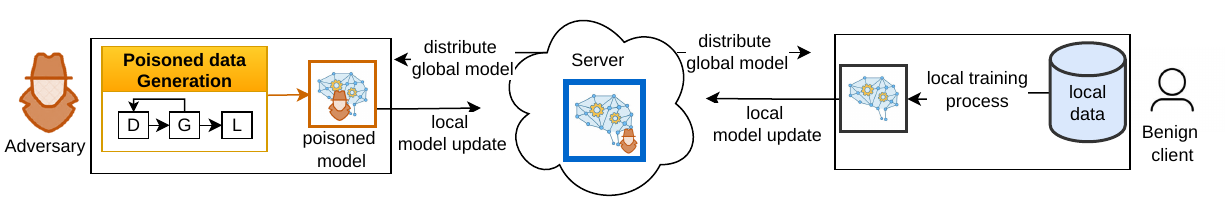}
   \vspace{-0.1in}
    \caption{GAN-based poisoning attack. The adversary trains a GAN to generate synthetic poisoned samples.}
    \Description{Synthetic sample generation using GAN for causing poisoning attack}
   \label{fig:gan}
\end{figure}

\begin{figure}[!htb]
    \centering
   \includegraphics[scale=0.85]{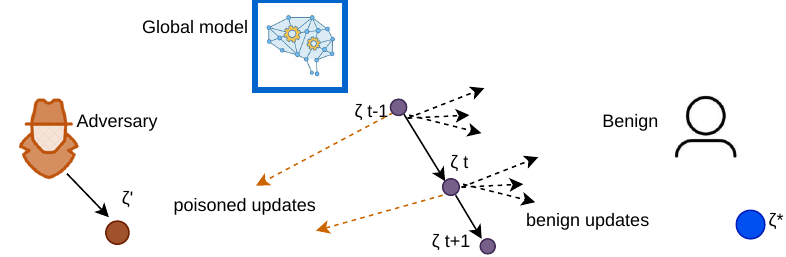}
    \caption{Illustrating model poisoning attack, where $\zeta^t$ denote the local model update in $t^{th}$ round. $\zeta^{*}$ represents the optimal model.}
    \Description{Illustration of model poisoning attack}
   \label{fig:replaceModel}
\end{figure}

As mentioned earlier, these attacks are caused by intentional poisoning and can be broadly divided into data or model poisoning. 
\noindent $\bullet$ {\textbf{\textit{Data poisoning attacks:}}}
Assuming that an attacker has access to and can modify the training data of multiple clients, data poisoning attacks become a significant concern. 
These attacks are further categorized by their nature as follows:
$(i)$ {\textit{Label-oriented attacks:}} They involve altering the labels of a subset of the training data~\cite{tolpegin2020data}. Specific labels are purposefully swapped (targeted), or labels are shuffled randomly (untargeted). Label flipping is a typical Dirty Label Poisoning (DLP) attack to create malicious gradients~\cite{xu2022rethinking}. A distance-aware attack in~\cite{sun2023attacking} selects the most vulnerable class for label flipping by calculating the mean feature vector of each class as:
\begin{math}
    \psi_l = \frac{1}{|D_{\text{a}}(l)|} \sum_{i \in D_{\text{a}}(l)} \Psi(i),
\end{math}
where $\psi_l$ denotes the mean of feature vectors in class $l$, $|D_{\text{a}}(l)|$ is the number of samples in the infected dataset, and $\Psi$ is the feature extraction function. 
To strengthen the attack, the least distant class is chosen based on the distances between classes:
$ D_m(l, l') = ||\psi_l - \psi_{l'}||_2. $
Another scenario is where the adversary cannot alter the training data labels due to a certification process that ensures label accuracy and demands that any changes be imperceptible (Clean-label poisoning (CLP)-induced attacks)~\cite{9835228, yang2023clean, peri2020deep, xia2023poisoning}. For instance, the samples are modified to render the poisoned instance ($m_s$) to resemble the base instance as:
\begin{math}
   m_s = \arg \min_x \left( \| \delta(i) - \delta(s) \|_2^2 + \eta \| i - s \|_2^2 \right), 
\end{math}
where the function $\delta(i)$ corresponds to the input $i$ propagating through the network, $s$ denotes the base instance, and $\eta$ controls the extent of this similarity to achieve the poisoning effect on the global model. 
Such attacks often leave the labels unaffected, making it more appealing and unnoticeable~\cite{yang2023clean},
but unfeasible in non-IID scenarios. 
$(ii)$  {\textit{Poisoned Samples Generation (PSG):}} Data poisoning attacks can also be carried out using Generative Adversarial Networks (GANs)~\cite{sun2024gan, alsereidi2024data, jere2020taxonomy}. Here, an attacker needs to train a GAN to replicate other participants' training samples~\cite{sun2024gan}, which are then used to craft scaled poisoning updates as their local updates, aiming to compromise the global model (Fig. \ref{fig:gan}). 
This method of attack is particularly effective and general since the generated data is realistic, making it difficult to detect and counter. 
In~\cite{zhang2020poisongan}, a poison data generation approach ($Data\_Gen$) is introduced that relies on iteratively updated global model parameters to regenerate samples for the targeted victims. 
Stepping beyond, the work~\cite{psychogyios2023gan} introduced a more sophisticated attack that can go unnoticed for several rounds. 
$(iii)$  {\textit{External Data Poisoning (EDP):}}
It resembles standard poisoning attacks but differs in that the poisoned samples originate from outside the original input distribution~\cite{ winkens2020contrastive, ren2019likelihood, sastry2020detecting}. 
They use data from entirely different domains with similar characteristics, or even random noise. Exploiting foreign data potentially impairs the model's integrity.
$(iv)$  {\textit{Colluding attack: }} This intense threat arises from the increased influence an attacker gains by controlling multiple clients within the FL framework~\cite{9798077}. They are mainly of two types, server-participant and participant-participant collusion~\cite{ranjan2022securing, lyu2023poisoning}, and can exploit the cooperative interactions to gain unauthorized insights or degrade model performance. 
A Sybil attack is a prominent example in which a single or small group of attackers creates multiple colluding identities to amplify the adversarial impact~\cite{tuor2021overcoming,cao2022flcert,sun2021data}.
\noindent $\bullet$ {\textbf{\textit{Model poisoning attacks:}}}
By targeting the manipulation of local model parameters or gradient updates directly, the attackers can deviate the model from the optimal update direction, as depicted in Fig. \ref{fig:replaceModel}. 
Attackers attempt to increase stealthiness by ensuring their model updates closely resemble benign updates \cite{shejwalkar2022back, zhou2021deep}. They might play random noise injection (RNI attack)~\cite{hossain2021desmp, fang2020local}, manipulate their local updates, or replace the global model with a malicious one (model replacement (MR attack))~\cite{sun2019can}, to disrupt the convergence of the global model, misclassify specific data samples, or increase energy consumption~\cite{cina2025energy}. Mathematically, a model poisoning attacker aims to replace the global model $\zeta_G^{(t+1)}$ with a malicious model $\tilde{\zeta}_G^{(t+1)}$.
To do so, when the global model converges such that $\sum_{j \in S_t} (\zeta_j^t - \zeta_G^t) \approx 0$, the attacker crafts its local model to cancel out the benign updates and steer the global model towards the malicious one. They may even scale their parameters to amplify their influence~\cite{9835228, bagdasaryan2020backdoor}.
Partial model replacement has been used by selfish clients to make the global model prioritize their local distribution~\cite{2026fairrfl}
Targeted Model Poisoning (TMP), where the adversary enhances their updates to have a significant impact on the global model while avoiding detection~\cite{bhagoji2019analyzing}, remains effective even against Byzantine-resilient aggregation methods. 
\begin{table*}[!htb]
\small
\centering
\vspace{0.1in}
\caption{Summary of recent poisoning attacks in FL. An attack is \textbf{Persistent} if it is carried out in every round.}
\vspace{-0.1in}
\resizebox{1\textwidth}{!}{\begin{tabular}{|>{\centering\arraybackslash}c|>{\centering\arraybackslash}m{1.5cm}|>{\centering\arraybackslash}m{4.6cm}|>{\centering\arraybackslash}m{1.2cm}|>{\centering\arraybackslash}m{1.4cm}|>{\centering\arraybackslash}m{1.7cm}|>{\centering\arraybackslash}m{2.8cm}|>{\centering\arraybackslash}m{2.0cm}|>{\centering\arraybackslash}m{2.3cm}|}
\hline
\rowcolor[HTML]{8c8c8c} \textbf{Paper} & \textbf{Threat} & \textbf{ Description} & \textbf{\begin{tabular}[c]{@{}c@{}}Non-IID \\ data\end{tabular}} & \multicolumn{1}{|c|}{\textbf{Persistent}}& \textbf{\begin{tabular}[c]{@{}c@{}}Similarity \\ space\end{tabular}} & \textbf{\begin{tabular}[c]{@{}c@{}}Evaluation \\ metrics\end{tabular}} & \multicolumn{1}{|c|}{\textbf{Dataset}} & \multicolumn{1}{|c|}{\textbf{Remarks}} 
\\ \hline
{\begin{tabular}[c]{@{}c@{}}\cite{peri2020deep}, \\ \cite{9835228},  \\ \cite{yang2023clean}, \\ \cite{xia2023poisoning}\end{tabular}}
& CLP attacks & Inject poison into samples but retain original labels, ensuring imperceptibility but embedding malicious traits.& $\times$&  $\checkmark$&   Negative Cosine Similarity (NCS) & Accuracy, Attack Success Rate (ASR), Peak Signal-to-Noise Ratio (PSNR) and Structural SIMilarity (SSIM) & MNIST, Fashion MNIST, CIFAR-10, ImageNet & Requires large-scale data to be effective \\ \hline
{\begin{tabular}[c]{@{}c@{}} \cite{xu2022rethinking},\\ \cite{tolpegin2020data} \end{tabular}} & DLP attacks & Introduce incorrect labels to data, creating misclassification risks upon model incorporation. &$\checkmark$& $\times$ & Euclidean Distance & Misclassification Rate, Attack Impact & MNIST, CIFAR-100, KDD, Amazon & More detectable than clean-label poisoning \\ \hline
{\begin{tabular}[c]{@{}c@{}} \cite{sun2024gan},  \cite{alsereidi2024data}, \\ \cite{jere2020taxonomy},  \cite{zhang2020poisongan}, \\ \cite{psychogyios2023gan}  \end{tabular}}& PSG with GANs & Use adversarially trained GANs to create poisoned samples that mimic real data, disrupting global model accuracy.  & $\checkmark$& $\checkmark$& Frechet Inception Distance, Wasserstein Distance&  Model Accuracy, Detection Rate & PlantVillage, CelebA, ImageNet, Tiny-ImageNet  &  Requires high computational resources \\ \hline
{\begin{tabular}[c]{@{}c@{}} \cite{sastry2020detecting}, \cite{winkens2020contrastive}, \\ \cite{ren2019likelihood}  \end{tabular}} & EDP attacks &  Insert samples from outside the original input distribution to impair model integrity.  &$\checkmark$& $\times$ & Jensen-Shannon Divergence (JSD) &  Model Deviation, ASR & OpenImages, MNIST, CIFAR-10 &  Limited by dataset shift detection methods \\ \hline
{\begin{tabular}[c]{@{}c@{}}  \cite{9798077}, \cite{ranjan2022securing}, \\ \cite{tuor2021overcoming}, \cite{cao2022flcert},  \\  \cite{sun2021data},\\ \cite{li2024contribution} \end{tabular}} & Colluding attacks & Multiple controlled clients collude to amplify attack effects, often seen in Sybil attacks. & $\checkmark$& $\checkmark$& Cosine Similarity& ASR, Model Deviation & MNIST, CIFAR-10 & Harder to execute with robust aggregation techniques \\ \hline
{\begin{tabular}[c]{@{}c@{}} \cite{hossain2021desmp}, \\ \cite{fang2020local} \end{tabular}} & RNI attacks & Inject random noise into gradients or models to disrupt model convergence.& $\times$& $\checkmark$& Kullback-Leibler (KL) divergence & Model Convergence, Error Rate & FEMNIST, MNIST, CIFAR-10 & Limited by robust gradient clipping methods \\ \hline
{\begin{tabular}[c]{@{}c@{}} \cite{sun2019can}, \\ \cite{9835228} \end{tabular}} & MR attacks & Replace the global model with a malicious version to ensure misclassification of specific instances. & $\checkmark$& $\checkmark$&  Euclidean Distance & Model Accuracy, ASR & ImageNet, MNIST, CIFAR-10 & Requires full control over model updates \\ \hline
{\begin{tabular}[c]{@{}c@{}} \cite{bhagoji2019analyzing} \\\cite{wang2020attack}  \end{tabular}}& TMP attacks & Adversary enhances their updates to create a significant impact on the global model while avoiding detection.& $\checkmark$& $\checkmark$&  KL divergence  & ASR, Stealth Score & MNIST, CIFAR-10 & Detection methods can limit effectiveness \\ \hline
 \cite{zhou2021deep} & OMP attacks & Use malicious neurons and regularization terms to poison or manipulate the objective function while mitigating catastrophic forgetting in NNs &$\times$& $\checkmark$& Wasserstein Distance & Loss Function Deviation, Model Accuracy & UCIHAR, MNIST, CIFAR-10 & Can be mitigated by adversarial training \\ \hline
{\begin{tabular}[c]{@{}c@{}} \cite{el2020genuinely}  \end{tabular}}& RRA attacks & Generate malicious gradients by reversing or randomly replacing benign gradients.  &  $\checkmark$& $\times$& Euclidean Distance & Model Divergence, ASR & FEMNIST, MNIST, CIFAR-10 & Countered by gradient aggregation methods \\ \hline
{\begin{tabular}[c]{@{}c@{}} \cite{wei2023covert}, \\ \cite{xie2020fall}  \end{tabular}} & DMP attacks & Optimize the distance between malicious and target models to evade detection. &$\checkmark$& $\checkmark$& Euclidean Distance & Euclidean Distance, Attack Stealth Score & FEMNIST, MNIST, CIFAR-100 & Requires precise attack control. \\ \hline
\end{tabular}
}
\label{tab:poisoning}
\end{table*}
Some Byzantine-robust methods (Krum, Trimmed-mean, and Median) can be compromised by directly manipulating local parameters even in a partial-knowledge setting~\cite{fang2020local}. An Optimization-based Model Poisoning (OMP) in~\cite{zhou2021deep} injects malicious neurons into the neural network's redundant space by leveraging the regularization term.
Reverse and Random attacks (RRA)~\cite{el2020genuinely} create adversarial gradients by reversing or randomly substituting the benign gradients. However, because the adversarial gradients deviate greatly from the benign ones, Byzantine-resilient FL methods can easily identify and eliminate them. 
Additionally, Dynamic Poisoning Methods (DMPs) that use flexible tactics to circumvent defenses~\cite {wei2023covert} and create malicious models using inner-product alteration~\cite{xie2020fall} also exist. 
 For quick reference, Table~\ref{tab:poisoning} delivers an insightful summary of major poisoning attacks in FL. 
\begin{table*}[!htb]
\small
\centering
\vspace{0.1in}
\caption{Summary of backdoor attacks and Free-riders participation in FL.}
\vspace{-0.1in}
\resizebox{1\textwidth}{!}{\begin{tabular}{|>{\centering\arraybackslash}c|>{\centering\arraybackslash}m{1.7cm}|>{\centering\arraybackslash}m{4.5cm}|>{\centering\arraybackslash}m{1.1cm}|>{\centering\arraybackslash}m{1.2cm}|>{\centering\arraybackslash}m{1.7cm}|>{\centering\arraybackslash}m{2.8cm}|>{\centering\arraybackslash}m{1.9cm}|>{\centering\arraybackslash}m{2.8cm}|}
\hline
\rowcolor[HTML]{8c8c8c} \textbf{Paper} & \textbf{Threat} & \textbf{Description} & \textbf{\begin{tabular}[c]{@{}c@{}}Non-IID \\ data\end{tabular}} & \multicolumn{1}{|c|}{\textbf{Persistence}}& \textbf{\begin{tabular}[c]{@{}c@{}}Similarity \\ space\end{tabular}} & \textbf{\begin{tabular}[c]{@{}c@{}}Evaluation \\ metrics\end{tabular}} & \multicolumn{1}{|c|}{\textbf{Dataset}} & \multicolumn{1}{|c|}{\textbf{Remarks}} 
\\ \hline
{\begin{tabular}[c]{@{}c@{}} \\ \cite{huynh2024combat} \end{tabular}}  & Single pattern attack & Adversarial clients inject the same malicious pattern into the model. & $\times$ & $\checkmark$ & Model weight similarity & ASR , Accuracy Drop & CIFAR-10, MNIST & Limited adaptability in diverse settings. \\ \hline
 {\begin{tabular}[c]{@{}c@{}} \cite{bagdasaryan2020backdoor},\\ \cite{abad2023sniper}\end{tabular}}  & Model weight modification & Modify weights to embed hidden patterns influencing specific tasks  while preserving accuracy. & $\checkmark$ & $\checkmark$ & Parameter space similarity & ASR, Accuracy & CIFAR-10, ImageNet & Hard to detect as attacked models resemble benign ones. \\ \hline
{\begin{tabular}[c]{@{}c@{}} \cite{xie2020dba}, \\ \cite{gong2022coordinated} \end{tabular}} & Multi-backdoor attack & Multiple coordinated adversaries inject diverse local patterns or segments of a global pattern.  & $\checkmark$ & $\checkmark$ & Model update clustering & ASR, False Positive Rate & CIFAR-10, Fashion-MNIST & High complexity makes detection challenging. \\ \hline 
{\begin{tabular}[c]{@{}c@{}} \cite{liu2020backdoor} \end{tabular}} & Feature-partioned attack & Backdoor embedded in feature-partitioned FL without label manipulation.   & $\checkmark$ & $\checkmark$ & Gradient similarity analysis & ASR, Accuracy & Purchase-100, FEMNIST & Gradient aggregation mitigates attack effectiveness. \\ \hline
{\begin{tabular}[c]{@{}c@{}} \cite{chen2020backdoor} \end{tabular}} & Federated meta-learning attack & Backdoor persists even after meta-training and fine-tuning on clean data.  & $\checkmark$ & $\checkmark$ & Task-specific weight similarity & ASR, Model Accuracy & Omniglot, Mini-ImageNet & Reducing fine-tuning effectiveness is a challenge. \\ \hline
{\begin{tabular}[c]{@{}c@{}} \cite{wang2020attack}  \end{tabular}} & Edge-case backdoor attack & Targeting underrepresented input data to induce misclassification.  & $\checkmark$ & $\checkmark$ & Adversarial loss detection & ASR, Robust Accuracy & CIFAR-10, GTSRB & Hard to detect due to its selective nature. \\ \hline
{\begin{tabular}[c]{@{}c@{}} \cite{mozaffari2024}, \\ \cite{cao2022mpaf}, \\  \cite{sagduyu2022free} \end{tabular}} & Free-riders & Global model without any valid contribution.  & $\times$ & $\times$ & Contribution similarity detection & Contribution Score, Model Performance Drop & MNIST, FEMNIST & Reduces overall model robustness. \\ \hline
{\begin{tabular}[c]{@{}c@{}} \cite{wang2022assessing}, \\ \cite{fraboni2021free} \end{tabular}} & Free-riders with noise & Introduce noise into the aggregation process & $\times$ & $\times$ & Update entropy analysis & Model Divergence, Accuracy Drop & CIFAR-10, ImageNet & Detection is difficult in low-noise cases. \\ \hline
{\begin{tabular}[c]{@{}c@{}} \cite{wan2022shielding} \end{tabular}} & Disguised small data attack & Training a model locally on a small dataset, pretending it is a large one, to gain incentives.   & $\checkmark$ & $\times$ & Dataset size estimation & Contribution Score, Model Accuracy & CIFAR-10, Shakespeare & Undermines fair performance \\ \hline
{\begin{tabular}[c]{@{}c@{}} \cite{pejo2023quality} \end{tabular}} & Random weights free-riders & Generate a gradient update matrix by randomly sampling parameters from a uniform distribution to appear benign. & $\times$ & $\times$ & Gradient distribution analysis & Update Similarity Score, Model Accuracy & MNIST, FEMNIST & Easily mitigated with aggregation rules. \\ \hline
\end{tabular}}
\vspace{-0.1in}
\label{tab:freeridingbackdoor}
\end{table*}
\subsubsection{Backdoor Attacks}
Adversaries perform backdoor attacks by injecting hidden trigger patterns into data to impair performance on particular class instances. Such attacks manifest during both training and inference phases~\cite{bagdasaryan2020backdoor}.
Their stealthy and selective behavior, which compromises model performance only when the trigger is active, has critical consequences for FL~\cite{wang2020attack}.
Detecting hidden triggers is highly complex, particularly in unexpected scenarios~\cite{bhagoji2019analyzing,wang2024invariant}.
Backdoor attacks can be classified into:
$(i)$ {\em Single-pattern attack:} All backdoor clients inject the same trigger, which makes their detection easier regardless of the attack's potency~\cite{ huynh2024combat}. The study in~\cite{bagdasaryan2020backdoor} attempted to alter model weights to create hidden functionalities to influence specific tasks without reducing main accuracy. Moving beyond simple data poisoning, joint data and model poisoning are employed~\cite{abad2023sniper, wang2020attack}. 
$(ii)$ {\em Coordinated attack:} Numerous coordinated adversaries injecting patterns or segments of a common pattern are typically harder to detect~\cite{xie2020dba} for the variation and spread of the embedded patterns. The study~\cite{xie2020dba} introduced a distributed variant where adversaries cooperate to embed various segments of a chosen trigger,
while~\cite{gong2022coordinated} considered the use of model-specific triggers. 
$(iii)$ {\em Evolving attacks:} The adversarial strength has evolved beyond typical scenarios to more intricate settings. In feature-partitioned FL~\cite{liu2020backdoor}, adversaries inject triggers without altering the label, although gradient aggregation helps to alleviate them. In federated meta-learning~\cite{chen2020backdoor}, backdoor attacks endure despite fine-tuning on clean data. Edge-case backdoors~\cite{wang2020attack} target sporadic samples, scaling parameters through Projected Gradient Descent (PGD).
\subsubsection{Free-riders}
Free-riders are deceptive participants who do not provide meaningful updates to the global model but rely on other participants to perform most of the training work~\cite{sagduyu2022free, cao2022mpaf, mozaffari2024, pejo2023quality}. 
They typically use only a limited portion of their dataset for training or introduce random noise or arbitrary model updates to conserve computational resources~\cite{sagduyu2022free, wang2022assessing, fraboni2021free, wan2022shielding}, ultimately compromising the overall model due to inferior data quality. 
A random-weights technique in~\cite{pejo2023quality} challenges detection by generating gradient updates via parameter sampling from a uniform distribution.
Table~\ref{tab:freeridingbackdoor} provides a summary discussion of free-riders and backdoor attacks.
\subsubsection{Channel Attack}
This section explores two FL channel attack scenarios. {\em First,} we examine susceptibilities and communication bottlenecks spawned from periodic updates and frequent interactions between the server and participants~\cite {ye2022decentralized}. 
An example is the man-in-the-middle (MITM) attack~\cite{wang2020man, vangala2022blockchain}. Since FL relies on the server to manage the learning process and aggregate updates, such attacks can intercept and alter communication, posing a critical vulnerability. The study in~\cite{yao2018two} highlighted how adversaries can exploit the training process by causing delays, cutting bandwidth, aggravating interference, and undermining model convergence.
{\em Second}, we discuss the risk of exploiting FL as a secret channel for stealthy interaction, where an intruder utilizes its features to enable concealed data exchanges, compromising the robustness.
More covert attacks ~\cite{hitaj2023fedcomm, costa2022turning}  can even turn the FL systems to covert channels of their desire. FedComm introduced in~\cite{hitaj2023fedcomm} encodes disguised messages into weights using advanced synchronization methods while adjusting updates to stay stealthy. 
\begin{table*}[!htb]
\vspace{0.1in}
\renewcommand{\arraystretch}{1}
\small
\centering
\caption{Summary of recent works on channel, DoS, and evasion attacks on FL frameworks.}
\vspace{-0.1in}
\resizebox{1\textwidth}{!}{\begin{tabular}{|>{\centering\arraybackslash}c|>{\centering\arraybackslash}m{1.7cm}|>{\centering\arraybackslash}m{3.6cm}|>{\centering\arraybackslash}m{1.2cm}|>{\centering\arraybackslash}m{1.4cm}|>{\centering\arraybackslash}m{2.2cm}|>{\centering\arraybackslash}m{2.8cm}|>{\centering\arraybackslash}m{2.2cm}|>{\centering\arraybackslash}m{2.5cm}|}
\hline
\rowcolor[HTML]{8c8c8c} \textbf{Paper} & \textbf{Threat} & \textbf{Description} & \textbf{\begin{tabular}[c]{@{}c@{}}Non-IID \\ data\end{tabular}} & \multicolumn{1}{|c|}{\textbf{Persistence}}& \textbf{\begin{tabular}[c]{@{}c@{}}Similarity \\ space\end{tabular}} & \textbf{\begin{tabular}[c]{@{}c@{}}Evaluation \\ metrics\end{tabular}} & \multicolumn{1}{|c|}{\textbf{Dataset}} & \multicolumn{1}{|c|}{\textbf{Remarks}} 
\\ \hline 
{\begin{tabular}[c]{@{}c@{}} \cite{wang2020man}, \\ \cite{vangala2022blockchain}  \end{tabular}} & MITM attack & Disrupt communication between the server and participants, creating a single point of failure. & $\checkmark$ & $\times$ & Communication delay & Packet loss, Communication time & CIFAR-10, MNIST & Exploits communication bottlenecks to intercept updates \\ \hline
{\begin{tabular}[c]{@{}c@{}}  \cite{yao2018two}  \end{tabular}} & Channel bottleneck exploitation & Exploit server-client communication bottlenecks by reducing bandwidth, introducing delays, and interference. & $\checkmark$ & $\checkmark$ & Latency, bandwidth reduction & Model accuracy, Convergence time &  CIFAR-10, MNIST & Can lead to severe performance degradation \\ \hline
 {\begin{tabular}[c]{@{}c@{}} \cite{hitaj2023fedcomm}  \end{tabular}}& Covert channel through advanced encoding & Embed hidden messages in model parameters using encoding techniques, utilizing weights and gradient updates. & $\checkmark$ & $\checkmark$ & Gradient modification, encoding & Model update consistency, Hidden data rate & WikiText-2, CIFAR-10, MNIST,  ESC-50  & Covert data transmission in model weights \\ \hline
 {\begin{tabular}[c]{@{}c@{}}  \cite{costa2022turning}  \end{tabular}}& Covert channel through input sample poisoning & Create a covert communication channel by encoding data bits into changes observed during federated training rounds.    & $\checkmark$ & $\checkmark$ & Data poisoning, input modification & Model accuracy, Detection time & MNIST, CIFAR-10 & Affects model training with undetectable input manipulation \\ \hline
{\begin{tabular}[c]{@{}c@{}} \cite{zhang2023denial} \\ \end{tabular}}  & DDoS attack & Target server and network traffic to prevent clients from connecting to the server.  & $\checkmark$ & $\checkmark$ & Network congestion, client disconnection & Connection uptime, Service disruption & EMNIST, MNIST, CIFAR-10 & Prevents system operation and model updates \\ \hline
{\begin{tabular}[c]{@{}c@{}} \cite{cao2022flcert}, \\ \cite{fung2020limitations} \end{tabular}} &  Training inflation (Sybil attacks) & Inflate malicious participants to disrupt the training process. & $\checkmark$ & $\checkmark$ & Identity falsification & Training effectiveness, Model accuracy & MNIST, CIFAR-10, Reddit, HAR,  KDDCup & Disrupt FL process with fake clients \\ \hline
{\begin{tabular}[c]{@{}c@{}} \cite{mothukuri2021survey}, \\ \cite{kurakin2017adversarial}  \end{tabular}} & White-box evasion attack & Use full access to model parameters to generate adversarial samples & $\checkmark$ & $\checkmark$ & Adversarial sample generation & Classification accuracy, Model robustness & MNIST, CIFAR-10, ImageNet & Misclassify samples using model knowledge \\ \hline
{\begin{tabular}[c]{@{}c@{}} \cite{chen2017zoo}  \end{tabular}} & Black-box evasion attack & Generate adversarial samples through interaction with the system.  & $\checkmark$ & $\checkmark$ & Optimization, boundary search & Misclassification rate, ASR & MNIST, CIFAR-10 & Attack without knowledge of model parameters \\ \hline
{\begin{tabular}[c]{@{}c@{}} \cite{kim2023characterizing}  \end{tabular}} &   Internal evasion attack & Manipulate internal training data to deceive the model and other clients  & $\checkmark$ & $\checkmark$ & Data manipulation, inference attacks & Classification accuracy, Model deviation & CIFAR-10, Fake News, Celeba & Affects inference phase \\ \hline
\end{tabular}
\label{table:others}
}
\vspace{-0.1in}
\end{table*}
\subsubsection{Denial-of-Service (DoS) Attack}
Traditional FL is vulnerable to DoS attacks, which substantially disrupt model convergence and affect output accuracy. Distributed DoS (DDoS) attacks targeting the FL servers, resource allocation, and network traffic can often go undetected~\cite{zhang2023denial, fung2020limitations}. Consequently, clients experience difficulty connecting to servers, disrupting communication and functionality. The study by~\cite{cao2022flcert} categorizes and labels untargeted attacks initiated by Sybils as DoS. 
The utilization of historical data in ~\cite{zhang2023denial} to iteratively tune the malicious model via neuron perturbations enhances DoS effectiveness. 
\subsubsection{Evasion Attack}
The adversary intentionally introduces minor malicious perturbations to input samples during the inference or deployment phase, leading the classifier to misclassify the samples with high probability~\cite{  kurakin2017adversarial}. They are categorized into white-box and black-box attacks based on the adversary's knowledge. 
In a white-box attack, the adversary has complete access to the learning algorithm and model parameters, allowing the formulation of adversarial samples using these details~\cite{mothukuri2021survey}.  
In contrast, the black-box attackers lack this information. They create adversarial samples through system interaction or by querying the system. 
A zeroth-order optimization~\cite{chen2017zoo} estimates model updates and generates adversarial samples.   
Internal evasion attacks, where the adversary clients execute an attack internally at test time to deceive other clients, are discussed in~\cite{kim2023characterizing}. 
At a glance, Table~\ref{table:others} summarizes recent works related to channel, DoS, and evasion attacks. 
\begin{table*}[!htb]
\vspace{0.1in}
\renewcommand{\arraystretch}{1}
\centering
\small
\caption{Summary of information stealing threats on FL systems.
}
\vspace{-0.1in}
\resizebox{1\textwidth}{!}{\begin{tabular}{|>{\centering\arraybackslash}c|>{\centering\arraybackslash}m{1.7cm}|>{\centering\arraybackslash}m{3.8cm}|>{\centering\arraybackslash}m{1.4cm}|>{\centering\arraybackslash}m{1.4cm}|>{\centering\arraybackslash}m{2.2cm}|>{\centering\arraybackslash}m{2.9cm}|>{\centering\arraybackslash}m{2.2cm}|>{\centering\arraybackslash}m{2.5cm}|}
\hline
\rowcolor[HTML]{8c8c8c} \textbf{Paper} & \textbf{Threat} & \textbf{Attack description} & \textbf{\begin{tabular}[c]{@{}c@{}}Non-IID \\ data\end{tabular}} & \multicolumn{1}{|c|}{\textbf{Persistence}}& \textbf{\begin{tabular}[c]{@{}c@{}}Simlarity \\ method\end{tabular}} & \textbf{\begin{tabular}[c]{@{}c@{}}Evaluation \\ metrics\end{tabular}} & \multicolumn{1}{|c|}{\textbf{Dataset}} & \multicolumn{1}{|c|}{\textbf{Remarks}} 
\\ \hline 
{\begin{tabular}[c]{@{}c@{}} \cite{he2024enhance} \end{tabular}} & Membership inference (Passive) & Infer whether a data point is a part of the training set by observing model updates. & $\checkmark$ & $\times$ & Cosine similarity & Accuracy of membership prediction & CIFAR-10, MNIST, FMNIST, CIFAR-100 & Can infer membership without active involvement. \\
\hline  
{\begin{tabular}[c]{@{}c@{}} \cite{gu2022cs} \end{tabular}} & Membership inference (Active) & Tamper the FL model to gain insights into the training data of other participants.   & $\checkmark$ & $\checkmark$ & Gradient ascent & Loss reduction rate, Accuracy of inference & CIFAR-10, MNIST & Active attack increases the risk by direct manipulation. \\ \hline
{\begin{tabular}[c]{@{}c@{}} \cite{10269696}, \\ \cite{9679121} \end{tabular}} & Source inference & Identify the source participant by analyzing model updates, exposing information about the data owner. & $\checkmark$ & $\checkmark$ & Gradient analysis & Identity recovery accuracy, Success rate of source identification & CIFAR-10, MNIST, CHMNIST & Risk of revealing the participant’s identity. \\ \hline
{\begin{tabular}[c]{@{}c@{}}  \cite{suri2022subject} \end{tabular}} & Subject inference  & Target data owners rather than records, aiming to infer their membership using a trained model.  & $\checkmark$ & $\checkmark$ & Loss analysis & Subject inference accuracy & MNIST, Fashion-MNIST & Requires model access post-training for the attack's effectiveness. \\
\hline 
{\begin{tabular}[c]{@{}c@{}}  \cite{zhang2020poisongan}   \end{tabular}} & Poisoning and inference & Combine poisoning techniques with inference methods to manipulate training data and infer information about participants.  & $\checkmark$ & $\checkmark$ & Gradient poisoning& Data manipulation accuracy, Inference accuracy & CIFAR-10, MNIST  & More potent when combined with poisoning attacks, allowing for model manipulation. \\
\hline 
{\begin{tabular}[c]{@{}c@{}}  \cite{arevalo2024task}, \\ \cite{mothukuri2021survey} \end{tabular}} & Attribute inference  & Target specific features of participants’ data to infer sensitive information without accessing the full data.  & $\checkmark$ & $\times$ & Gradient analysis & Attribute inference accuracy & CIFAR-10, MNIST  & Challenging to detect the attacker's presence. \\
\hline
{\begin{tabular}[c]{@{}c@{}} \cite{chen2021robust} \end{tabular}} & Property inference  & Uncover private properties of participants using the global model, even with secure aggregation in place.   & $\checkmark$ & $\checkmark$ & Model aggregation analysis& Property recovery accuracy, Performance degradation & CIFAR-10, MNIST & Potential to escape even a secure aggregation. \\ \hline
{\begin{tabular}[c]{@{}c@{}} \cite{song2020analyzing} \end{tabular}}  & GAN-based inference (Passive) & Analyze user inputs at the server level to infer private training data. & $\checkmark$ & $\times$ & GAN-based model updates & Data recovery accuracy & MNIST  & Relies on passive observation, making it harder to detect. \\ \hline
{\begin{tabular}[c]{@{}c@{}}\cite{zhang2024nspfl} \end{tabular}} & GAN-based inference (Active) & Attacker sends global updates to a specific client to infer their training data.  & $\checkmark$ & $\checkmark$ & GAN-based model updates & Data inference success, Attack effectiveness & MNIST, Synthetic & Active manipulation increases the attack's effectiveness. \\
\hline
{\begin{tabular}[c]{@{}c@{}}\cite{10621090} \end{tabular}} & Stealthy label inference  & Bypass secure aggregation to recover private labels from clients using highly accurate fishing models.  &  $\checkmark$ & $\checkmark$ & Gradient extraction models & Label inference accuracy, Attack success rate & CIFAR-10, MNIST  & Can accurately infer private labels. \\ \hline
{\begin{tabular}[c]{@{}c@{}} \cite{li2023model},  \cite{9945997},  \\ \cite{10376357}  \end{tabular}} & Model extraction  & Replicate the target model by querying it and analyzing predictions; watermarking to claim ownership. & $\checkmark$ & $\times$ & Cosine similarity, Euclidean distance & ASR, Model similarity & MNIST, CIFAR-100 & High success rate but requires multiple queries. \\
\hline
{\begin{tabular}[c]{@{}c@{}}\cite{qiu2024hashvfl},  \cite{chen2023privacy}, \\ \cite{loo2023understanding} \end{tabular}} & \multirow{2}{*}{\begin{tabular}[c]{@{}c@{}} Reconstruc-\\ tion \end{tabular}}  & Recreate samples from model parameters & $\checkmark$ & $\checkmark$ &SSIM, PSNR, NCS & Reconstruction accuracy, Privacy leakage & CIFAR-10, FER, MNIST, CBPD, CRITEO, IMBd  & More effective on highly predictive models \\
\hline
{\begin{tabular}[c]{@{}c@{}} \cite{wang2024privacy} \cite{liu2023mgia}, \\ \cite{sun2024gi},  \cite{10621090},  \\ \cite{gupta2022long},     \cite{geiping2020inverting},\\  \cite{issa2024rve}  \end{tabular}}& Inversion  & Reconstruct sensitive data from gradients. & $\checkmark$ & $\checkmark$ & Mean Squared Error (MSE), Cosine Similarity, PSNR & Data recovery success, Feature inference accuracy & CIFAR-100, STL-10, FMNIST, Flower images & Pose greater risk due to exposure of multiple clients' data \\
\hline
{\begin{tabular}[c]{@{}c@{}}  \cite{xu2021else}  \end{tabular}} & {\begin{tabular}[c]{@{}c@{}} Eavesdrop-\\ ping \end{tabular}}  & Passive attack to infer sensitive information. & $\checkmark$ & $\times$ & JSD, KL Divergence & Amount of leaked data, Inference success rate & MNIST, LEAF synthetic & Hard to detect due to passive nature \\
\hline
\end{tabular}
\label{table:inf1}
}
\vspace{-0.1in}
\end{table*}
\subsection{Information Stealing Threats}
While FL is designed to prevent direct data sharing, research has shown that gradient exchanges can expose sensitive details, such as class representations and data membership, to attackers (both passive and active)~\cite{moriai2019privacy, 9945997, 9524709, 10184496}.  Since models reflect high-level data statistics, attackers can exploit them to infer private information. In extreme scenarios, attackers might even recover actual training samples or extract the labels solely from the update gradients.
\subsubsection{Inference Attacks}
\begin{figure}[!htb]
    \centering
   \includegraphics[width=0.9\linewidth]{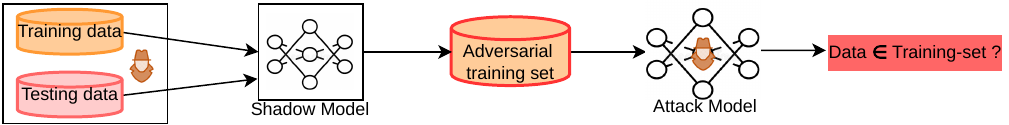}
    \caption{In a membership inference attack, through a shadow model, the adversary can infer whether a specific data sample was part of the training set by analyzing the model outputs. } 
    \Description{membership inference attack example}
   \label{fig:mem_inf}
\end{figure}

During the training stage, the model can unintentionally expose sensitive private information or user data~\cite{zhang2024nspfl, gu2022cs, 10269696, suri2022subject}. An honest but curious server or user may infer information on the training data without prior knowledge~\cite{10184496, 9679121}. Inference attacks can be classified into four major types:
$(i)$ \textit{Membership inference:} Attackers aim to determine whether a specific data sample has been used in the training. The attack can either be active or passive~\cite{he2024enhance} and typically leverages a shadow model to mimic the target model's behavior, enabling the attacker to analyze the model's outputs and infer membership status, as illustrated in Fig.~\ref{fig:mem_inf}.
In a passive attack, the adversary observes updated model parameters and infers membership without altering the learning process. 
In contrast, an active attack directly tampers with model training to gain more insights into the data of other participants. 
The attacker can apply a gradient ascent attack, in which it checks the loss over subsequent communication rounds in FL to determine whether the sample is likely part of the training set~\cite{gu2022cs}. 
$(ii)$ \textit{Source inference:} This attack goes beyond general membership inference by focusing on identifying the specific FL participants.
The exposure of source information can indeed raise substantial concerns about information protection and confidentiality~\cite{10269696, 9679121}. 
The work~\cite{suri2022subject} introduced two novel black-box attacks that identify multiple participants collectively, without requiring access to the model's parameters at every training round.
$(iii)$ \textit{Attribute inference:} This attack attempts to infer particular attributes or properties that participants do not intend to disclose, focusing on dataset subsets rather than the entire collection~\cite{arevalo2024task,mothukuri2021survey}. Unlike general attributes associated with the main task, these properties are specific to individual participants, making such attacks effective even if secure aggregation techniques are employed. 
Attackers can analyze the global model to uncover such attributes without affecting the overall performance on the primary task~\cite{chen2021robust}. 
$(iv)$ \textit{Label inference:}
 With a capability to bypass secure aggregation and recover private labels post-aggregation~\cite{10621090}, label inference is stronger than other inference-based attacks.
Client-specific fishing models extract client gradients, enabling large-scale label inference with 100\% accuracy, emphasizing the increased risk from leaked gradients.
Further, GAN-based inference attacks\cite{song2020analyzing} have also been explored in both passive and active modes. 
Passive mode enables an attacker to analyze user inputs at the server level, whereas active mode involves sending global updates to a specific client. 
According to~\cite{song2020analyzing}, even with a small percentage of gradients, attackers can infer training data.
\subsubsection{Model Extraction Attacks}
This attack captures model predictions by querying the trained model with diverse inputs, enabling attackers to recreate a functionally equivalent model. Through iterative refinement, the recreated model's outputs are adjusted to better match the original model's predictions.
While clients typically have model access, attackers may attempt to steal models to claim ownership of the aggregated model~\cite{9945997}.
Further, an internal adversary~\cite{10376357} can also play an attack aiming to deprive other participants of their model ownership rights, enabling the attacker to deceptively claim the model as their intellectual property. 
Direct analysis can reconstruct models with high precision, focusing on critical points where derivatives are mostly zero~\cite{li2023model}.
\subsubsection{Reconstruction and Inversion Attacks}
In these attacks, an adversary aims to reconstruct training samples from a learning model's parameters, across both black-box and white-box scenarios~\cite{qiu2024hashvfl, chen2023privacy}.
Within the context of inference task distribution, a malicious client can reconstruct random inputs without direct access to the data or computations of other participants~\cite{chen2023privacy}. This threat scales with model architecture, as demonstrated by the potential to recover entire training sets in infinite-width models~\cite{loo2023understanding}. 
Even with moderate DP, the attacker can still infer stored data, although increased privacy controls lower accuracy~\cite{mothukuri2021survey, zhang2023survey}.  
Importantly, strong predictive frameworks are more vulnerable because they form tight connections across data features and labels. 
Participant-side attacks can masquerade as typical FL participants while stealthily recreating the data, posing a major harm.
Inversion techniques, specifically gradient and model inversion, represent a critical subset of these threats that occur at both the client and server levels~\cite{liu2023mgia, 10621090, gupta2022long}. 
In a gradient inversion attack, an adversary intercepts shared gradients and uses an auxiliary model to iteratively create data samples that mirror the client's actual data~\cite{sun2024gi}. 
These attacks differ significantly in extent depending on the adversary's position.
Server-side attacks pose a systemic threat by potentially exposing data from all the participants~\cite{geiping2020inverting}.
Instead, client-side model inversion infers features of another client's data by observing model outputs and iteratively adjusting inputs to match observed outputs~\cite{issa2024rve, wang2024privacy}.
\subsubsection{Eavesdropping}
Eavesdropping is an attack in which adversaries exploit inadequate client security to intercept and monitor client-server interactions in FL.  With this, attackers can obtain confidential details about clients and their engagements. 
These assaults generally go unnoticed as they involve a passive approach or re-encryption tricks that mask the break-in.
Although normally untargeted, it can be targeted by an intruder seeking specific information~\cite{xu2021else}. 
A summary of different information-stealing threats is provided in Table~\ref{table:inf1}. 
 We present the key insights into the above-mentioned threats to robustness through the following aspects: type of threat, its target, vulnerable component of FL, and threat complexity, in Table~\ref{tab:attacks_in_fl}.
 \begin{table}[!htb]
\centering
\vspace{0.1in}
\caption{Insights into threats to robustness in FL.}
\vspace{-0.1in}
\resizebox{0.8\linewidth}{!}{
\begin{threeparttable}
\begin{tabular}{|m{4.5cm}|m{3.8cm}|c|c|c|c|c|c|c|}
\hline
\rowcolor[HTML]{8c8c8c} \textbf{Attacks} & \textbf{Attack Target} & \multicolumn{5}{c|}{\textbf{Vulnerable Component}} & \textbf{Complexity} & \textbf{Type} \\ 
\cline{3-7}
\rowcolor[HTML]{8c8c8c} & & \textbf{C} & \textbf{S} & \textbf{M} & \textbf{D} & \textbf{Comm} & & \\ 
\hline
Clean-Label Poisoning & Training Data & $\checkmark$ & & & & & H & A \\ 
\hline
Dirty-Label Poisoning & Training Data & $\checkmark$ & & & & & M & A \\ 
\hline
Poisoned Samples Generation& Training Data & $\checkmark$ & & & & & H & A \\ 
\hline
Model Poisoning & Global Model & $\checkmark$ & $\checkmark$ & & & & H & A \\ 
\hline
Backdoor Attacks & Global Model & $\checkmark$ & & & & & H & A \\ 
\hline
Evasion Attacks & Model Prediction & $\checkmark$ & & $\checkmark$ & & & M & A \\ 
\hline
Training Rules Manipulation & Learning Process & $\checkmark$ & & & $\checkmark$ & & H & A \\ 
\hline
Free-Riding Attacks & FL Participation & & & & $\checkmark$ & & L & P \\ 
\hline
Inference Attacks & Private Data & & $\checkmark$ & & & & M & P \\ 
\hline
GAN Reconstruction & Private Data & $\checkmark$ & & & & $\checkmark$ & H & P \\ 
\hline
Non-Robust Aggregation & Aggregation Process & & & & $\checkmark$ & & M & P \\ 
\hline
Malicious Server & Global Model & & $\checkmark$ & & & & H & A \\ 
\hline
Model Replacement Attack & Global Model & $\checkmark$ & $\checkmark$ & & & & H & A \\ 
\hline
Model Inversion Attack & Private Data & & $\checkmark$ & $\checkmark$ & & & H & P \\ 
\hline
Model Extraction Attack & Model Parameters & & $\checkmark$ & $\checkmark$ & & & M & P \\ 
\hline
Model Ownership Attack & Model Ownership & & $\checkmark$ & $\checkmark$ & & & M & A \\ 
\hline
FL Channel attacks & FL Process & & & & $\checkmark$ & $\checkmark$ & M & P \\ 
\hline
Man-in-the-Middle Attacks & Model Updates & & & & & $\checkmark$ & H & A \\ 
\hline
Sybil Attacks & FL Integrity & $\checkmark$ & & & & & H & A \\ 
\hline
Eavesdropping Attacks & Communication Data & & & & & $\checkmark$ & M & P \\  \hline
\end{tabular}
\begin{tablenotes}
\footnotesize \item C: Client, S: Server, M: Model, D: Distributed FL nature, Comm: Communication, L: Low, M: Medium, H: High, A: Active, and P: Passive attacks
\end{tablenotes}
\end{threeparttable}
}
\label{tab:attacks_in_fl}
\vspace{-0.2in}
\end{table}
\section{Robust Aggregation Strategies} \label{sec:aggr}
Safeguarding the integrity of the aggregation process is imperative, particularly when faced with potentially corrupted model updates from participating clients of the FL framework~\cite{li2019rsa, andrew2021differentially}. 
FL aggregation algorithms are still in the early phases of development, and their robustness is an area demanding active ongoing research and advancement~\cite{chen2023privacy, liu2022privacy, du2023efficient}. 
In a decentralized FL scheme, the key issue lies in filtering out anomalous or malicious contributions while retaining the authenticity of legitimate contributions during aggregation~\cite{wan2022shielding}. To solve this, several solutions have been developed, attempting to reinforce the aggregation phase and enhance the robustness of FL frameworks~\cite{zhang2026secure, wang2024turbosvm, pejo2023quality}.  \\
This section critically reviews robust aggregation strategies that enhance the resilience and effectiveness of combining locally trained models into a cohesive global model, and later introduces a taxonomy to classify them.
\subsection{Aggregation Robustness Challenges}
Models can be combined using either parameter-based or output-based methods. The parameter-based approach is the most common. It brings together trainable parameters, like weights or gradients, from local models~\cite{9833835}. The output-based approach uses representations from the models, such as output logits or compressed sketches. For example, Fedmask~\cite{10571602} helps mobile devices with limited computing power by having them learn binary masks, which are then combined by the server. 
Studies suggest that incorporating cryptographic techniques such as  Homomorphic Encryption (HE), which enables computations over encrypted data, to safeguard sensitive information, and DP, which adds random noise to outputs, can enhance robustness against information-stealing threats~\cite{andrew2021differentially}. The assumption followed in centralized FL that the server is reliable yet curious, and the clients are trustworthy, is easily undermined in the presence of adversaries.
Recent developments in robust aggregation place more emphasis on enabling clients to confirm that the server has completed the aggregation correctly, in addition to safeguarding local updates~\cite{brunetta2021non}. 
Decentralized aggregation solutions, such as gossip mechanisms and blockchain technology, are advised to eliminate single points of failure by removing the need for a central server. 
Other options are centered on optimizing contract mechanisms, employing robust stochastic model aggregation, and organizing small committees to successfully handle model updates~\cite{li2019rsa}. Solutions such as Trusted execution environments (TEE) and Multi-party computation (MPC) further reveal the ongoing efforts to boost security and robustness of FL frameworks~\cite{sotthiwat2021partially, zhao2021sear, kadhe2020fastsecagg, zhang2021shufflefl}.
On the other hand, other vital aspects, such as enhancing training quality and handling heterogeneity, remain underexplored. The efficacy and scalability in large-scale FL frameworks also demand deeper investigation~\cite{kuznetsov2021securefl}. 
\begin{figure}[!htb]
    \centering
   \includegraphics[width=0.9\linewidth]{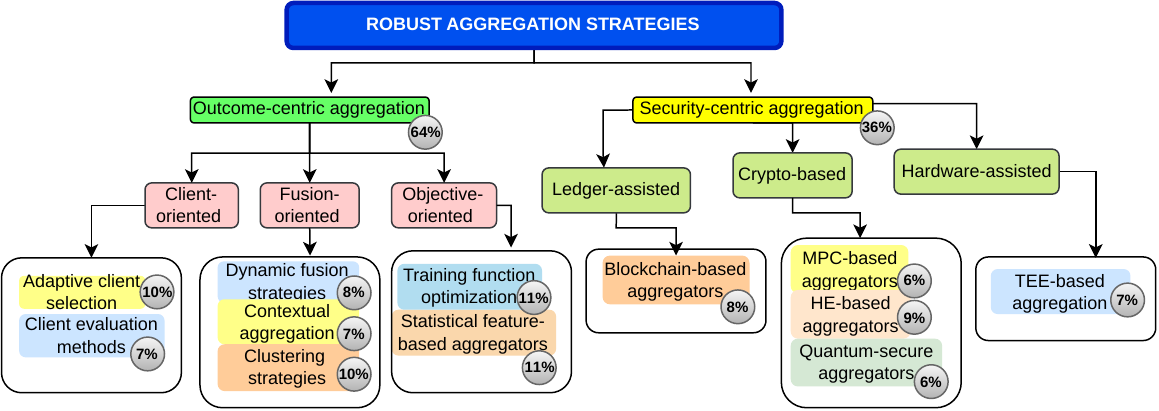}
   \vspace{-0.1in}
    \caption{Taxonomy of the robust aggregation strategies along with the relative proportion (\%) of reviewed publications.}
    \Description{Robust aggregation strategies taxonomy}
   \label{fig:taxaggre}
   \vspace{-0.1in}
\end{figure}
\subsection{Proposed Taxonomy of Aggregation Strategies} \label{subsec:aggr_tax}
This section proposes a new taxonomy, shown in Fig.~\ref{fig:taxaggre}, for analyzing and classifying robust aggregation solutions. They are divided into two macro-categories according to their key design focus:
(i) \textbf{outcome-centric}: enhances robustness by boosting the learning outcome of the global model under potentially falsified client updates, and (ii) \textbf{security-centric}: reinforces robustness by safeguarding the integrity, verifiability, and confidentiality of the aggregation phase itself. 
\subsubsection{Outcome-centric Aggregation} This category includes client-oriented methods, which enhance robustness by selectively prioritizing high-quality and trustworthy clients through adaptive selection and client evaluation strategies; fusion-oriented methods redesign the way local models are merged via dynamic fusion, contextual aggregation, and clustering-based strategies; and objective-oriented methods, which mold the training objective using statistical feature-based aggregators and customized loss or regularization designs to lessen attackers' impact.
{\textit{(i) Adaptive selection of clients:}} 
To enhance robustness during aggregation~\cite{du2023efficient}, it is crucial to prioritize participants with better communication abilities, higher prediction outcomes, or models that better align with the global model. 
Clients are chosen in frameworks such as Eiffel~\cite{sultana2022eiffel} based on update frequency, processing ability, and data size. 
The robustness of FL framework is further strengthened by sophisticated solutions like  Lyapunov optimization~\cite{perazzone2025communication}, sequential Kalman filters~\cite{yan2025correlated}, and bandit learning methods~\cite{wan2022shielding}, which use clients likely to deliver high-quality updates and alleviate bad contributions. 
In~\cite{wang2024turbosvm}, an SVM-based aggregation process is introduced for lazy clients in cross-device FL, accelerating convergence without burdening clients. 
{\textit{(ii) Client evaluation mechanisms:}} Discovering the appropriate weight per client and recognizing adversaries can be accomplished by evaluating the local updates. Maintaining a server-specific model~\cite{cao2021fltrust} is one innovative approach. Before the start of the collaborative training process, the server uses a carefully selected mini dataset to train a baseline server-version of the model. The cosine similarity between client updates and the server-version is assessed to understand how well they align. The higher score updates are given more weight.
 In a different study~\cite{park2021sageflow}, weights assignment to clients followed an approach based on the entropy of each gradient received on an evaluation dataset. 
{\textit{(iii) Dynamic fusion approaches:}}  
A reliable way to reduce training times without compromising model accuracy is dynamic local model fusion approaches~\cite{lee2021adaptive, wu2023hiflash, sun2022asynchronous}.
The issue of low-accuracy models caused by traditional fully asynchronous aggregation has been addressed by methods using client numbers or interval time windows. The trade-off between model performance and training time can be effectively balanced by dynamically selecting the number of aggregations per round~\cite{lee2021adaptive}. 
To enhance robustness, techniques for handling stale models have also been suggested, such as avoiding out-of-date local models from the aggregation process~\cite{wu2023hiflash} and temporal weight decay methods~\cite{zhang2021dynamic}.  
{\textit{(iv) Contextual model aggregation:}} By adapting the model aggregation process to fit a particular context, contextual aggregation methods solve slow convergence issues~\cite{liu2022privacy, nguyen2022contextual}. The study~\cite{nguyen2022contextual} sought to ensure consistent loss reduction at each optimization round by imposing a context-based bound on loss reduction specific to the participants and assuming smoothness of the overall loss function. The results demonstrated significant gains in robustness and convergence speed, outperforming conventional techniques.
{\textit{(v) Clustering-based solutions:}} In hierarchical FL, clustering has emerged as a robust strategy to improve both model performance and resource usage, solving key challenges related to non-IID data and communication efficacy~\cite{xiao2022efl}. Clients with similar characteristics or data distributions are grouped into clusters to enable effective model aggregation within each cluster. This improves the robustness of the FL process. Different clustering-based strategies, including those that use communication capabilities and device-to-device interactions within clusters, have been investigated~\cite{lin2021semi}.
One such framework is the FedSim~\cite{palihawadana2022fedsim}, which employs k-means clustering to guide the aggregation process. This approach is further extended using diverse distance metrics, such as Manhattan and cosine distances~\cite{wang2021adaptive} and hybrid metrics~\cite{augello2023dcfl}.
Beyond distance metrics, studies have explored clustering in terms of data distribution to support the training of personalized models, designed for specific environmental contexts~\cite{li2022hpflcn}. This approach can be extended using data-similarity-based client partitioning approaches to promote collaboration and disincentivize selfish behavior.
{\textit{(vi) Statistical-feature-based aggregators:}} To prioritize the gradients from the most reliable clients, some studies formulate statistical-feature-based aggregators that protect the global model from potentially poisonous updates~\cite{10571602}. 
Some of the existing approaches, such as Trimmed Mean, Krum, Median, and Multi-Krum~\cite{peng2024mean, wang2024invariant}, use robust estimators either individually or jointly to mitigate the influence of outliers and adversarial updates.
For instance, by employing the geometric median to aggregate local updates via a Weiszfeld-type algorithm, the study in~\cite{9721118} introduced a robust aggregation solution that safeguards against compromised participants without revealing local contributions and performs well under severe corruption. 
{\textit{(vii) Training function optimization:}} Researchers have examined approaches for optimizing the distributed training loss function to improve robustness, intending to refine the learning process itself.  Certain studies have introduced appealing techniques~\cite{li2019rsa, andrew2021differentially, zhao2024huber}, though they are still in the early stages of development.
For example, ~\cite{li2019rsa} applies regularization to the loss function to reduce the local model deviations from the global model during training, thereby enhancing the resilience of the learning process. Moreover, ~\cite{andrew2021differentially} presented a dynamic clipping value estimated online to adapt to various situations. In ~\cite{zhao2024huber}, an extension of the Huber loss function is provided to devise an optimal loss structure, delivering theoretical guarantees in IID and slightly heterogeneous scenarios.
\subsubsection{Security-centric Aggregation}
\begin{figure}[!htb]
    \centering
   \includegraphics[scale=0.8]{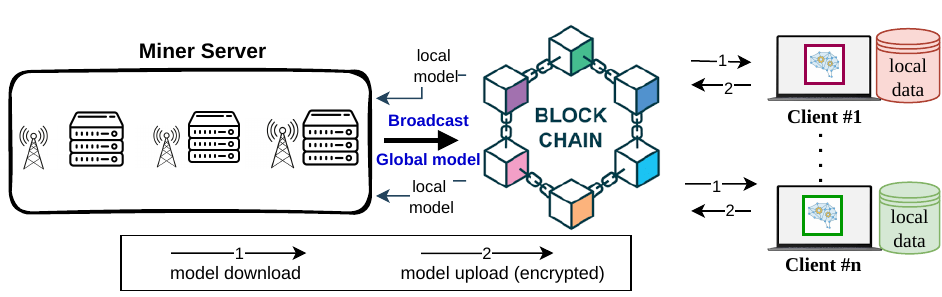}
    \caption{Blockchain-based FL: encrypted updates and aggregated results are safely recorded; miner servers perform model aggregation.}
    \Description{blockchain-based FL for aggregation robustness}
   \label{fig:blockchain}
\end{figure}

 This group includes ledger-based methods that employ decentralized models such as blockchain to provide tamper-resistant logging, reputation, and auditability for the aggregation process,  hardware-assisted methods that employ TEEs to perform aggregation within safe enclaves, and cryptographical techniques such as MPC, HE, and Quantum Secure Aggregation (QSA) that safeguard the confidentiality and verifiability of model updates and aggregation outcomes. They are reviewed as follows.
{\textit{(i) MPC-based aggregators:}} Secure Multi-Party Computation has been researched to improve both privacy and robustness in federated aggregation~\cite{sotthiwat2021partially}. 
These approaches reduce the chances of information leakage by enabling the global model to be computed without disclosing specific local models.
While the studies such as~\cite{brunetta2021non} prioritize verifiable sharing schemes and a dual-stage aggregation, where a committee is initially selected for aggregating the models, the research in~\cite{kadhe2020fastsecagg} utilizes Fast Fourier Transform (FFT) based confidential sharing as a substitute to the standard Shamir confidential sharing method. By lowering the risk of one-point failure and boosting security, sharing secret-shared local updates over several aggregation servers can also strengthen robustness.
{\textit{(ii) HE-based aggregators:}} 
FL frameworks that rely on HE-based aggregators typically adopt a shared public key-based encryption approach to protect the local models, allowing the server to use the additive homomorphic property of the encryption system to aggregate the encrypted models, without decrypting them~\cite{xu2020privacy, 9524709}. For the system to be reliable and robust~\cite{liu2022privacy}, the secret key management is vital. Three main secret key handling choices are covered in the literature, and each has a distinct influence on robustness and privacy in FL systems~\cite{zhu2021distributed}. 
The three key handling practices are discussed as follows: 
(1) Secret key sharing among users --  this approach distributes the secret key to all users but kept concealed from the server. Each participant has access to the global model. Even though this configuration permits aggregation using cryptographic systems like Paillier, RSA, lattice-based,  ElGamal, and BGN~\cite{fang2021privacy}, its robustness is constrained by potential flaws in key dissemination.
(2) Central server holds the key -- only the server holds the secret key, protecting the privacy of the global model. However, because the server can decrypt encrypted models, this centralization can undermine robustness and risk client model privacy. Additional strategies, such as model masking or using a trusted party to handle the secret key, are required to improve robustness and privacy. 
(3) Threshold-based key handling -- this enhances the shared key management by mandating a certain number of users to collaborate to perform decryption utilizing techniques such as  ElGamal and threshold Paillier~\cite{zhu2021distributed}. 
\begin{table*}[!htb]
\vspace{0.1in}
\renewcommand{\arraystretch}{1}
\centering
\caption{Summary of robust aggregation approaches.}  \label{tab:agg} 
\vspace{-0.1in}
\resizebox{1\textwidth}{!}{
\begin{tabular}{|m{2.3cm}|m{1.6cm}|m{5.5cm}|m{5.5cm}|m{6.0cm}|m{5.5cm}|}
\hline
\rowcolor[HTML]{8c8c8c} \textbf{Category} & \textbf{Paper} & \textbf{Methods} & \textbf{Key Features} & \textbf{Advantages} & \textbf{Drawbacks}  \\
\hline
Client model evaluation & \begin{tabular}[c]{@{}c@{}}  \cite{cao2021fltrust},  \cite{9721118}, \\ \cite{park2021sageflow} \end{tabular} & Cosine similarity based evaluation of gradients, Geometric median aggregation, Entropy-based client weighting. & Prioritizes gradients with higher cosine similarity, Robust aggregation via geometric median, Personalization via entropy-based weighting. &  Enhances robustness against malicious clients, Reduces impact of outliers. & Computational overhead in gradient similarity calculations, Requires additional validation metrics, May not generalize to all non-IID settings. \\
\hline
Quantum secure aggregation & \begin{tabular}[c]{@{}c@{}}  \cite{chehimi2022quantum},  \cite{javeed2024quantum}, \\ \cite{yang2022post},   \cite{zhang2022federated} \end{tabular} & Quantum bit representation of model parameters, Entangled qubits for model aggregation, Post-quantum secure protocol using homomorphic pseudorandom generator. & Secure aggregation using qubits, Low computational complexity, Post-quantum security, Compatible with different model architectures. & High resilience against attacks, Ensures privacy without sacrificing efficiency. & Requires specialized quantum hardware, Limited scalability in classical systems, Not widely adopted due to early-stage development. \\
\hline
Dynamic fusion of local models & \begin{tabular}[c]{@{}c@{}}  \cite{lee2021adaptive},  \cite{wu2023hiflash}, \\ \cite{sun2022asynchronous} \end{tabular} & Dynamic fusion of local models based on time windows, Adaptive aggregation strategies, Temporal weight decay strategies. & Excludes stale models, Implements deadline-based aggregation. & Handles stale model updates effectively, Improves aggregation efficiency. & Performance depends on accurate time-window selection, High communication overhead for frequent updates, Increased computation for temporal decay management. \\
\hline
Statistical features & \begin{tabular}[c]{@{}c@{}}  \cite{peng2024mean},  \\ \cite{wang2024invariant} \end{tabular} & Trimmed Mean, Median-Krum, Multi-Krum aggregation. & Median-Krum joint method, Robust aggregation under adversarial conditions, Effective accuracy and convergence under malicious conditions. & Robust against adversarial manipulations, Convergence under malicious attacks, Effective under non-IID settings & May discard useful model updates, computationally expensive, May not perform well in extreme data heterogeneity scenarios. \\
\hline
Adaptive client selection & \begin{tabular}[c]{@{}c@{}}  \cite{du2023efficient},  \cite{sultana2022eiffel}, \\ \cite{wan2022shielding} \end{tabular} & Multi-armed bandit strategy, Gradient update norms, Radial-basis functions, Adaptive client selection based on communication capacity. & Multi-armed bandit approach, Dynamic client evaluation, Balancing exploration vs. exploitation, Focus on gradient norms and model alignment. & Dynamic selection based on exploration-exploitation trade-off, Efficient client evaluation to reduce communication overhead. & Selection bias may impact global model accuracy.  \\
\hline
Training function optimization & \begin{tabular}[c]{@{}c@{}}  \cite{li2019rsa},  \cite{andrew2021differentially}, \\ \cite{zhao2024huber} \end{tabular} & Regularization in loss function, Dynamic clipping value, Huber loss function. & Loss function regularization, Dynamic adaptation for non-IID scenarios, Huber loss extension for optimal robustness. & Adaptive to non-IID scenarios, Improves robustness of convergence. & Requires careful tuning of regularization parameters. \\
\hline
Contextual model aggregation & \begin{tabular}[c]{@{}c@{}}  \cite{liu2022privacy},  \cite{nguyen2022contextual} \end{tabular} & Context-dependent aggregation bounds, Smooth loss function assumption & Context-dependent aggregation, Smoothness assumption for loss functions, Robust optimization per device context. & Optimizes aggregation based on device context. & Complexity increases with diverse contexts. \\
\hline
MPC-based approaches & \begin{tabular}[c]{@{}c@{}}  \cite{sotthiwat2021partially},  \cite{brunetta2021non}, \\ \cite{kadhe2020fastsecagg},  \cite{xu2020privacy} \end{tabular} & Fast Fourier Transform-based secret sharing, Two-step aggregation process, Verifiable sharing schemes & Fast Fourier Transform for secret sharing, Verifiable MPC schemes, Use of multiple servers for improved security and privacy. & Improved privacy via MPC, Supports collaborative computation, Resilience against single point of failure. & High computational and communication costs, Requires careful protocol design, Limited efficiency for large models. \\
\hline
Blockchain-based approaches & \begin{tabular}[c]{@{}c@{}}  \cite{chen2021robust},  \cite{vangala2022blockchain}, \\ \cite{hamouda2023ppss},  \cite{chen2024credible} \end{tabular} & Reputation-based reward systems, Blockchain for secure model aggregation, Model encryption during upload & Decentralized aggregation, Reputation-based model evaluation, Ensure model integrity and prevent tampering using blockchain. & Ensures transparency and traceability, Prevents model manipulation. & Slower convergence due to blockchain verification, High communication and storage overhead. \\
\hline
TEE-based approaches & \begin{tabular}[c]{@{}c@{}}  \cite{zhao2021sear},  \cite{zhang2021shufflefl}, \\ \cite{kuznetsov2021securefl} \end{tabular} & Trust execution environment for secure aggregation, DP techniques with TEEs, Model shuffling for additional security. & Secure model aggregation via TEEs, DP-enhanced privacy, Randomization via model shuffling, ML in TEEs for enhanced security. & Efficient computation, lower communication overhead, stronger protection against insider attacks. & Hardware dependency and trust assumption on hardware vendors, Memory and execution constraints, Vulnerable to side-channel attacks. \\
\hline
HE approaches & \begin{tabular}[c]{@{}c@{}}  \cite{9524709},  \cite{liu2022privacy}, \\ \cite{zhu2021distributed},  \cite{fang2021privacy} \end{tabular} & Additive HE, Secret key management methods (shared, centralized, threshold-based). & Privacy-preserving aggregation, various key management strategies, Additive HE for model aggregation. & Strong privacy guarantees, Compatible with different FL frameworks & Increased computational overhead with large-scale models, Demands careful key management. \\
\hline
Clustering-based approaches & \begin{tabular}[c]{@{}c@{}}  \cite{xiao2022efl},  \cite{lin2021semi}, \\ \cite{palihawadana2022fedsim},  \cite{wang2021adaptive}, \\ \cite{li2022hpflcn}, \cite{augello2023dcfl} \end{tabular} & K-means clustering, Client similarity metrics (Manhattan, cosine distances), Data distribution-based clustering. & Improved aggregation efficiency, Targeted model aggregation within clusters, Enhanced collaboration among similar clients. & Reduces aggregation overhead, Improves accuracy over diverse datasets. & Performance drops with incorrect clustering, requires accurate clustering criteria. \\
\hline
\end{tabular}
}
\vspace{-0.1in}
\end{table*}
{\textit{(iii) Quantum secure aggregation:}}
Quantum Secure Aggregation (QSA) schemes are receiving popularity~\cite{chehimi2022quantum, javeed2024quantum, yang2022post} since they can be used to strengthen robustness against evolving challenges. By encoding local model parameters in quantum bits ({\em qubits}), QSA can boost security and efficiency in aggregating them~\cite{zhang2022federated}, guaranteeing strong resilience against exposure to semi-honest adversaries. Eavesdropping can be discovered and eliminated by QSA's robust security settings. A 3-round post-QSA mechanism is devised by~\cite{yang2022post} to deal with quantum threat contexts. It employs an additive homomorphic decryption that relies on Shamir secret sharing and a homomorphic pseudorandom generator based on single-masking to guarantee robust functionality even when participants drop out. This mechanism is a promising strategy for reliable and robust aggregation in quantum-inspired FL since it exhibits impressive run-time efficiency and preserves privacy under a semi-honest threat model.
{\textit{(iv) Blockchain-based aggregators:}}
Blockchain technology, with its decentralized and tamper-resistant features, is increasingly being utilized to enhance FL robustness and security~\cite{chen2021robust, vangala2022blockchain, hamouda2023ppss}. 
In such a setup, clients download the global model from the blockchain, then train and encrypt their models before uploading them back~\cite{chen2024credible, kalapaaking2022blockchain}. 
Miner servers aggregate these models and update the global model, which is then distributed to other nodes (See Fig.~\ref{fig:blockchain}).
Recent developments include reputation-based reward systems where each aggregation node evaluates and reports the quality of local models to the blockchain. Integrating reputation with blockchain's inherent security helps enhance FL robustness and quality by validating and rewarding model contributions based on performance and reputation~\cite{ranathunga2022blockchain}.
{\textit{(v) TEE-based aggregation:}} 
The TEEs provide a hardware-isolated enclave within a processor, shielding sensitive code and data from the potentially compromised host operating system or Rich Execution Environment (REE)~\cite{zhao2021sear, zhang2021shufflefl}. 
In a robust FL framework, participants transmit encrypted local models to the REE, which serves as a gateway to the TEE, which securely decrypts and aggregates these updates, ensuring that raw model parameters remain inaccessible to the central server before the finalized global update is returned to the REE for distribution~\cite{zhao2021sear}.
Furthermore, executing the entire training process within the TEE environment has emerged as a comprehensive strategy to maintain end-to-end confidentiality of both algorithmic logic and data throughout the lifecycle of the aggregation process~\cite{kuznetsov2021securefl}. 
Table~\ref{tab:agg} provides a quick view on robust aggregators focusing on key features, advantages, and drawbacks.
\begin{table*}[!htb]
\centering
\vspace{0.1in}
\caption{Key insights on robust aggregation strategies.} \label{tab:insightaggregation}
\vspace{-0.1in}
\renewcommand*{\arraystretch}{1}
\resizebox{1\textwidth}{!}{
\begin{tabular}{|m{2cm}|m{4cm}|m{4cm}|m{4cm}|m{3cm}|m{3cm}|m{5cm}|}
\hline
\rowcolor[HTML]{8c8c8c} \textbf{Key Challenge} & \textbf{Underlying Causes} & \textbf{Impact on Aggregation} & \textbf{Existing Solutions} & \textbf{Evaluation Metrics} & \textbf{Trade-offs Involved} & \textbf{Gaps in Research} \\
\hline
Statistical heterogeneity & Non-IID data distribution, Device diversity, Varying client participation,  Variability in local datasets. & Divergence in local models affects global model convergence and performance. & Bayesian non-parametric methods, Neuron matching, Probabilistic federated neural matching. & Accuracy, Model convergence rate, Gradient similarity. & Increased computation for better adaptability requires careful tuning. & Generalizing methods for complex neural networks, Adaptive sampling, Meta-learning, Federated transfer learning, Hybrid approaches for improved convergence. \\ \hline
Fairness and bias mitigation & Demographic biases in models, Disproportionate impact of non-IID data. &  Biased model updates, reducing overall model fairness and generalization. & FairFL, Adaptive sampling, Bias-aware aggregation strategies. & Model fairness metrics (e.g., demographic parity, equal opportunity). & Increased computational costs and require additional fairness constraints in optimization. & Advances in Federated debiasing techniques, Fairness-aware aggregation.  \\ \hline
Bottlenecks in communication & High number of clients, Limited bandwidth and connectivity constraints, Frequent communication rounds, Latency Constraints, Communication Overhead.  & Slows down model updates and convergence & AirComp-based FL (Over-the-air computation), Intelligent reflective surfaces (IRS), Multi-relay techniques, Gradient sparsification, Quantization, Local update compression, 6G integration. & Latency, Communication overhead, Model convergence speed. & Accuracy loss, Higher infrastructure cost. & 6G for higher efficiency and scalability, AI-driven network optimization, Edge caching. \\ \hline
Secure aggregation  & Model poisoning, Inference attacks, Sybil attacks, Data poisoning, Backdoor attacks. &  Risk of poisoning and inference attacks compromise FL security. & Robust aggregation, Anomaly detection TEE, Cryptographic filtering, Blockchain. & Attack resistance, Privacy preservation, Adversarial robustness. & Trade-offs in model utility, Higher computational and storage costs for improved security. & Enhancing blockchain security for decentralized FL, Secure multi-party computation (SMC), Decentralized consensus mechanisms.
  \\ \hline 
 Robustness and efficiency tradeoff & High computational and communication costs, Resource-constrained edge devices.  & Balancing security, accuracy, and resource constraints, Delay in model convergence. & Adaptive aggregation techniques, Sparsification and quantization, Cryptographic solutions.  & Accuracy loss, Model convergence time, Computational cost, Computational overhead.  & Model robustness and computational overhead, Security, and aggregation speed.  & Dynamic aggregation methods that optimize performance under varying conditions, Trade-off-aware federated optimization strategies, lightweight security mechanisms, and robustness under real-time FL conditions.  \\ \hline 
 Quantum aggregation & Instability due to quantum errors, Quantum noise, error correction, and limited availability of quantum hardware.  & Faster aggregation, Higher security via quantum cryptographic techniques.  & Quantum secure multi-party computation, Quantum-enhanced HE, Hybrid quantum-classical FL aggregation models. & Aggregation speed, Computational complexity, Noise resilience in quantum operations, Robustness against quantum attacks. & Quantum and classical processing, Stability and computational speed, Scalability. & Stable quantum aggregation, PQC- methods, Error mitigation techniques, Hybrid quantum-classical models.  \\ \hline 
\end{tabular}
}
\vspace{-0.2in}
\end{table*}
\subsection{Key Insights on Robust Aggregators}
Despite progress in robust aggregation, several challenges persist and remain crucial to the continued advancement of FL systems.
{\em Statistical heterogeneity} remains a major challenge due to non-IID data, affecting global model convergence. 
{\em Fairness and bias} mitigation demands attention towards addressing demographic imbalances in FL through bias-aware aggregation approaches with reduced computational costs. {\em Communication challenges} arising from limited bandwidth and high client numbers hinder model convergence, demanding future improvements in technologies such as over-the-air computation and gradient sparsification.  
The use of blockchain-based solutions and cryptographic verifications for decentralized FL has strengthened the aggregation robustness against inference and poisoning assaults. 
 While adaptive and dynamic techniques improve aggregation performance, reliability, and accuracy, resource limits must be balanced to ensure the {\em trade-off between robustness and efficiency}. More efficient and secure aggregation provided by QSA makes it a viable future path. However, hardware constraints and quantum noise lead to stability and scalability challenges. 
Table~\ref{tab:insightaggregation} presents the key insights for a quick understanding of the research gaps in the robust aggregation methodologies.
The table summarizes the root causes, impacts on aggregation, reported solutions, assessment approaches, trade-offs, and research gaps for each challenge, rendering a thorough summary of the current literature to support future plans for boosting aggregation robustness in FL.
\section{Robustness via Defensive Strategies} \label{sec:def}
Given FL's susceptibility to diverse threats, robust defensive strategies are critical to protect local updates and aggregation results against established and evolving threats.
\noindent $\bullet$ {\textbf{\textit{Limitations of conventional defenses:}}}  Classical centralized defensive measures, such as anomaly detection and robust loss functions, rely on direct inspection of training data or regulating  participants~\cite{jiang2023complement, 9518221}. Such approaches, however, are not directly viable to FL, as the centralized server has confined capabilities to monitor the local updates or parameters during the training stage~\cite{zhao2024huber, 9833835}. Local updates generated by the participants are vulnerable to different challenges, demanding early-phase security measures. Protecting training data is equally important, as it supports the development of robust models~\cite{yu2023untargeted, 10589557, wang2023sparsfa}. 
\begin{figure}[!htb]
    \centering
    \includegraphics[width=0.95\linewidth]{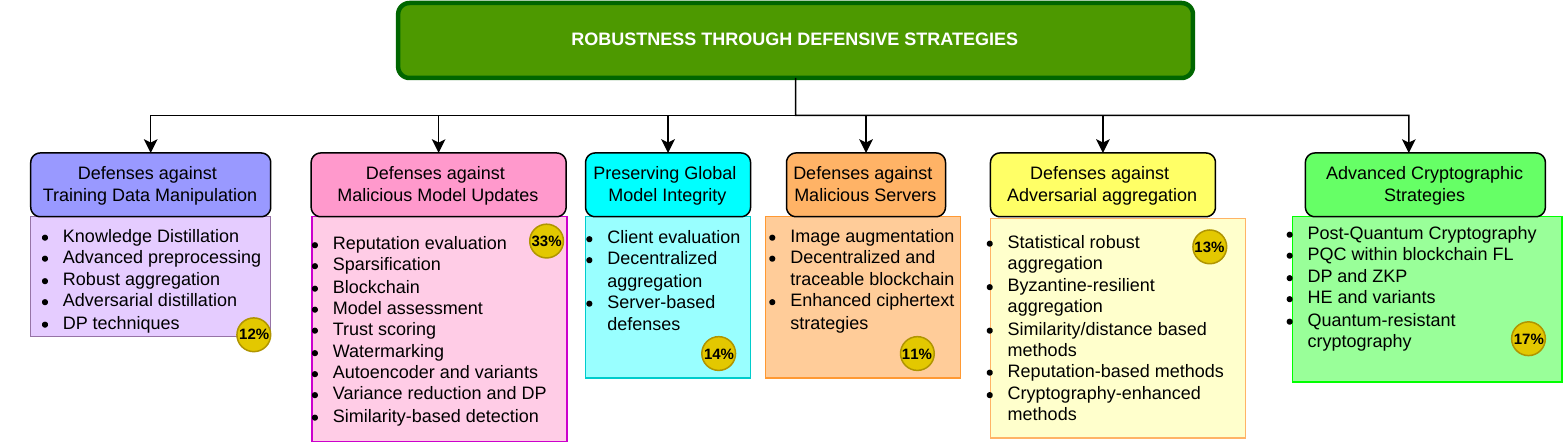}
    \caption{Taxonomy: Robustness through defensive strategies along with the relative proportion (\%) of reviewed publications.}
    \Description{Taxonomy of defensive strategies for upholding robustness}
   \label{fig:taxrobust}
\end{figure}
\subsection{Proposed Taxonomy of Defensive Strategies} With the growing complexity of adversarial threats, new defensive measures are evolving to thwart the attacks~\cite{9945997, liu2020backdoor}. 
The protection of client data and the integrity of local updates are guaranteed by client-level protective measures~\cite{Wan_2023}. New vulnerabilities could appear during the aggregation stage, when the server combines local model updates. Aggregated updates could expose personal information or tilt the global model if rigorous security measures aren't enforced~\cite{fung2020limitations, 9721118, fraboni2021free, li2019rsa}. For this reason, strategic countermeasures are vital. 
In order to guarantee the availability, confidentiality, and integrity of model components and training data, a thorough methodology that takes into consideration both local and global levels is important~\cite{9887909}. 
In contrast to current surveys, which categorize defenses by attack type, we prefer a more comprehensive strategy because many mechanisms are known to handle multiple attacks. 
Fig.~\ref{fig:taxrobust} gives the proposed taxonomy of defensive strategies for achieving robustness against hostile threats.
\subsubsection{Defenses against Training Data Manipulation}
Advanced preprocessing approaches to safeguard training data have been investigated recently~\cite{xia2023poisoning}. This strengthens data security by preventing data poisoning and property inference attacks~\cite{gupta2023performance}. 
An interesting example is provided in~\cite{xu2024flpm}, which employs an enhanced variational autoencoder (VAE) for property split and property variance management to circumvent inference attacks while maintaining model accuracy. This strategy successfully reduces the success rates of poisoning and inference attacks, advancing FL security.
In addition, \textit{DP techniques} are utilized to protect confidential data, further enhancing the security measures available for FL systems~\cite{andrew2021differentially}. Empirical evidence reveals that these methods can successfully mitigate the potency of reconstruction attacks.
The significance of \textit{Knowledge distillation (KD)} in boosting security and robustness of FL is highlighted in recent studies. The utility of global KD for retrieving useful knowledge while removing malicious inputs from infected clients is discussed in~\cite{park2023}. \textit{Adversarial distillation} as a defense mechanism is also investigated to enhance the robustness of the FL model against backdoor attacks by altering the distillation process during federated communication~\cite{zhu2023adfl}. 
These advancements demonstrate the effectiveness and adaptability in addressing privacy concerns, client-side attacks, and backdoors, promoting the field of FL toward stronger security and robustness~\cite{9524709}. 
\subsubsection{Defenses against Malicious Model Updates}
Robust aggregation, which does not necessitate the explicit identification of hostile clients, is considered a common defensive measure against a variety of poisoning attacks~\cite{peng2024mean, wang2024invariant, 9721118}.
However, the most successful defenses against malicious model updates include a variety of methods intended to identify and reduce detrimental effects from hostile clients~\cite{mozaffari2024, ma2022privacy}.  A common approach is \text{similarity-based detection}, which assumes that benign updates, though unique, will typically align a particular reference model or pattern~\cite{cao2021fltrust, wang2021adaptive}.  Malicious updates, on the other hand, are likely to break such assumptions.  To find deviations, each update is compared to an aggregated model or a known best model, using criteria such as Euclidean distance or cosine similarity. Local models are assessed using \textit{similarity-based recognition and trust scoring}, which assign ratings based on validation datasets or similarity factors. Clients exhibiting consistent, benign-like updates gain more weights via adaptive mechanisms, whereas distrustful updates are scaled-down or omitted from the aggregation process~\cite{cao2021fltrust, wang2021adaptive, wang2020attack, zhang2023survey, zhang2024nspfl}. 
When an attack has already contaminated the global model, mitigation measures prioritize identification of the parameter space where the attack's long-term impact resides and perturbing it during the training process~\cite{sun2021fl}. With this, residual effects can be neutralized to provide robustness against such risks. Moreover, the use of single-masking approaches helps to conceal actual client data, while cloud-stored data inhibits malevolent participants from initiating DoS behaviors, boosting FL system resilience~\cite{zhang2024nspfl}. One of the main drawbacks of such methods is that they often need extra datasets or TEEs to verify the integrity of updates, which is not always practical. 
Furthermore, these tactics can often tolerate only a certain number of malicious users and may not be robust against multiple types of poisoning scenarios~\cite{9887909, zhang2021shufflefl}. The paper~\cite{9887909} addresses this concern using a truth-discovering method that defends against multiple poisoning attacks without needing additional datasets, even when the proportion of adversaries exceeds half of the total users.
Other sophisticated methods to develop robust FL frameworks include \textit{Variance reduction and DP} methods~\cite{zhang2023byzantine, jiang2023complement, 10589557}. These enhancements strengthen security while maintaining a better balance between utility, privacy, and efficiency.
In addition to enabling effective preprocessing, \textit{autoencoder-based solutions} have evolved as powerful defenses against deceptive model updates. With the ability to learn compressed representations, autoencoders can effectively distinguish benign from deceptive updates, eliminating corrupted contributions prior to aggregation~\cite{xu2024flpm}. 
\textit{Sparsification} as a defense mechanism also exhibits its excellence in both boosting communication efficacy and robust security~\cite{jiang2023complement, 9518221, 10589557, wang2023sparsfa}. For instance, a peer-to-peer FL approach in~\cite{wang2023sparsfa} optimizes performance while providing robust defenses against multiple attacks. 
\textit{Watermarking} has also been employed as a defense strategy to ensure model integrity and ownership by including distinctive patterns during the training phase~\cite{Watermarking2024}.
Watermarks can be revealed to establish ownership and stop unauthorized duplication of the model, despite the fact that they remain hidden during regular operations. 
But preserving model functionality and privacy while maintaining watermark robustness is still difficult~\cite{xu2024robwerobustwatermarkembedding}.
In addition, blockchain technology provides options to use smart contracts for model integrity verification, guaranteeing that submitted models fulfill specifications~\cite{ranathunga2022blockchain, chen2021robust, chen2024credible, kalapaaking2022blockchain}. 
\begin{table*}[!htb]
\centering
\vspace{0.1in}
\caption{Summary on robustness via defensive strategies.} \label{tab:defenses-summary}
\vspace{-0.1in}
\renewcommand*{\arraystretch}{1}
\resizebox{1\textwidth}{!}{
\begin{tabular}{|m{1.8cm}|m{2.2cm}|m{6cm}|m{4.5cm}|m{3.2cm}|m{2.1cm}|m{6cm}|m{6cm}|}
\hline
\rowcolor[HTML]{8c8c8c} \textbf{Category} & \textbf{Paper} & \textbf{Methods} & \textbf{Key features} & \textbf{Evaluation metrics} & \textbf{Datasets} & \textbf{Advantages} & \textbf{Drawbacks} \\
\hline
Against training-data manipulation
&
\begin{tabular}[c]{@{}c@{}} \cite{xia2023poisoning},\cite{xu2024flpm}, \\ \cite{andrew2021differentially},\cite{park2023}, \\ \cite{zhu2023adfl} \end{tabular}
&
Advanced preprocessing, local data filtering, enhanced VAE for property division or variation control, DP on data or gradients, global KD to filter malicious client knowledge, adversarial distillation against backdoors.
&
Reduces poisoning and inference risk, provides noise-based privacy, and filters poisoned samples before aggregation.
&
ASR, property inference AUC, reconstruction success, and clean accuracy.
&
{\begin{tabular}[c]{@{}c@{}}  MNIST, \\ CIFAR-10/100,  \\ EMNIST \end{tabular}} 
&
Protects data and intermediate representations, reduces attack surface before aggregation, preserves utility via KD/distillation.
&
DP/KD/VAEs add overhead, utility loss if over-tuned, need auxiliary data, hyperparameter sensitivity. 
\\
\hline
Against malicious model updates
&
\begin{tabular}[c]{@{}c@{}}
\vspace{-1em}\\ \cite{wang2024invariant} \cite{wang2021adaptive},  \cite{cao2021fltrust}, \\    \cite{peng2024mean},  \cite{wang2020attack},  \cite{Watermarking2024},\\  \cite{zhang2021shufflefl},   \cite{zhang2024nspfl},  \cite{chen2024credible}, \\  \cite{kalapaaking2022blockchain}, \cite{ranathunga2022blockchain}, \cite{9518221}, \\ \cite{10589557}, \cite{wang2023sparsfa}, \cite{jiang2023complement}, \\ \cite{xu2024flpm}, \cite{9887909}   \cite{xu2024robwerobustwatermarkembedding}, \\ \cite{sun2021fl},\cite{ma2022privacy}, \cite{zhang2023byzantine} \\   \\  \end{tabular}
&
Robust aggregation rules, similarity or trust-based filtering, truth-discovery, HE-based, variance-reduced + DP, autoencoder-based anomaly filtering, sparsification, watermarking, blockchain/smart-contracts.
&
Server or client-side update inspection, down-weight or drop suspicious gradients, and cryptographic traceability for updates.
&
Accuracy under attack, ASR, detection precision/recall, convergence, watermark detection rate, and blockchain latency.
&
{\begin{tabular}[c]{@{}c@{}}  MNIST, \\ CIFAR-10/100, \\ EMNIST, \\  FEMNIST, \\ synthetic datasets\end{tabular}} 
&
Broad coverage of model-poisoning and byzantine behaviors, tolerate high malicious ratios, can recover from polluted rounds; reduces single point of failure, integrity, and ownership guarantees.
&
Bounded assumption on adversarial strength, demand extra validation data/ TEE, autoencoders, AEs/watermarks/blockchain overheads, sensitivity to non-IID, and hyperparameter tuning. 
\\
\hline
Preserving global model integrity
&
\begin{tabular}[c]{@{}c@{}}  \cite{sultana2022eiffel},\cite{liu2022privacy}, \\ \cite{du2023efficient},\cite{cao2022flcert}, \\ \cite{uprety2021mitigating},\cite{andreina2021baffle}, \\ \cite{issa2024rve} \end{tabular}
&
Global integrity assessment across training rounds, decentralized multi-model aggregation with voting, trust scores via global / client validation, and privacy-aware and personalized encoders.
&
Post-aggregation checking, detect and neutralize persistent poisoning effects, majority voting across models, maintain privacy while preserving utility.
&
Global accuracy under strong poisoning; integrity or trust scores; fraction of compromised updates detected and discarded; privacy-utility trade-off metrics.
&
{\begin{tabular}[c]{@{}c@{}}   EMNIST, \\ FEMNIST, \\ MNIST, \\ CIFAR-10, \\ synthetic datasets. \end{tabular}} 
&
Adds a global sanity check layer beyond local defenses, can recover from polluted rounds, and reduces a single point of failure via multiple globals.
&
Depends on representative validation sets/clients, higher coordination/computation overhead, and personalized encoders have extra complexity.
\\
\hline
Against malicious servers
&
\begin{tabular}[c]{@{}c@{}} \vspace{-3em}\\ \cite{10184496}, \cite{mothukuri2021survey},  \\ \cite{Jeter2023OASIS},\cite{10571602}, \\
 \cite{han2024privacy}, \cite{hao2023robust}, \\
\cite{guo2023fast},\cite{zhou2020pirate}, \\
\cite{ma2022federated},\cite{ye2022decentralized}, \\
\cite{9833835},\cite{tang2024flexible}  \\  \end{tabular}
&
Server-side threat analyses, image augmentation, enhanced ciphertext schemes, decentralized / blockchain FL, auditing and monitoring, limit or detect client identification by the server.
&
Limit gradient inversion, server-side reconstruction, client identification, protect against server model manipulation, and accountability for server operations using decentralization and auditing.
&
Privacy leakage metrics (e.g., inversion success), identification success, global accuracy under a malicious server, audit, and consensus performance.
&
{\begin{tabular}[c]{@{}c@{}}  CIFAR-10/100, \\ MNIST, \\ synthetic datasets.\end{tabular}} 
&
Targets the strongest adversary (the server) explicitly, preserves utility, and increases transparency and accountability.
&
Blockchain/auditing introduces latency, storage, and system complexity, tailored to specific modalities (e.g., images),  and cryptography demands careful key and protocol management.
\\
\hline
Adversarial-resilient aggregation
& \begin{tabular}[c]{@{}c@{}} Refer \\ section \ref{subsec:aggr_tax} \end{tabular}
& Statistical robust aggregation, byzantine-resilient aggregation, similarity/distance-based methods, reputation-based methods, cryptography-enhanced methods 
& 
Tolerate adversarial or corrupted gradients, detect and suppress abnormal updates and outliers during aggregation, account for reliable participants, protect update confidentiality and integrity.
& 
Global accuracy under attack, ASR, Byzantine tolerance level, convergence stability, computation and communication overhead.
& 
\begin{tabular}[c]{@{}c@{}} MNIST, \\ CIFAR-10/100, \\ EMNIST, \\ FEMNIST \\ ImageNet \\ \qquad \end{tabular}
&
Enhances robustness against adversarial behaviors, tolerates a fraction of malicious clients, enhances stability and reliability of global model updates.
& 
Assumes bounded adversaries, performance may get affected under highly non-IID data,  similarity and reputation mechanisms introduce additional computation, cryptography-enhanced methods increase overhead.
\\ 
\hline
Advanced cryptographic strategies
&
\begin{tabular}[c]{@{}c@{}} \vspace{-2em}\\ \cite{xia2023poisoning},\cite{sotthiwat2021partially}, \\ \cite{zhu2021distributed},\cite{ma2024vpfl}, \\ \cite{javeed2024quantum},\cite{chehimi2022quantum}, \\ \cite{gurung2023performance},\cite{ma2022shieldfl}, \\ \cite{moriai2019privacy},\cite{moshawrab2023polyflag_svm} \\  \end{tabular}
&
HE, partial HE, two-trapdoor HE with secure cosine similarity, ZKPs for verifiable training, PQC, quantum-resistant signatures + blockchain, PE for SVM-based FL.
&
Guarantee confidentiality and integrity of updates and aggregation,  public verifiability, and quantum-resistant security.
&
Byzantine tolerance level, runtime and memory overhead, proof generation/verification time/size, comm. cost, task accuracy under encrypted/PQC/PE.
&
{\begin{tabular}[c]{@{}c@{}} MNIST, \\ CIFAR-10, \\ synthetic datasets.\end{tabular}} 
&
Offer strong, provable security guarantees, can be combined with other defenses, higher tolerance in heterogeneous environments.
&
Higher overhead, especially for deep models, polynomial approximations needed for deep non-linear models, complex key management, large-scale deployment challenges.
\\
\hline
\end{tabular}
}
\end{table*}
\subsubsection{Preserving Global Model Integrity}
Evaluating the integrity and performance of the updated global model is essential to improve FL's robustness to model poisoning threats~\cite{sultana2022eiffel}.
Despite the existence of many robust aggregation techniques, these measures may not fully guarantee FL robustness in the face of highly potent adversaries.
As a result, even in the absence of further attacks, the negative consequences induced by the attack can persist through the later rounds if the global model is poisoned~\cite{andrew2021differentially, liu2022privacy, du2023efficient}.
Additionally, the framework is susceptible to adversarial manipulation due to its reliance on a single global model. This has been tackled with the emergence of \textit{decentralized aggregation schemes}, where clients collectively train the global model, each of which contributes to the ultimate prediction through majority voting~\cite{cao2022flcert}. Although it is still tricky to recognize specific attackers, this approach increases robustness against poisoning attempts. 
Devising effective \textit{trust measures for clients} is vital for accurate assessment.
A global validation set can evaluate the global model after each client submission, updating client trustworthiness based on model performance~\cite{uprety2021mitigating}.
 However, the effectiveness of this defense depends on the quality of the test set, prompting the exploration of alternatives.
 For e.g., instead of centralized testing, a modified global model can be evaluated by selecting validation clients using their local data~\cite{andreina2021baffle}. If flagged as compromised by a majority of these clients, the server discards the update, ensuring the global integrity and reinforcing overall robustness.
~\cite{issa2024rve} focuses on mitigating such threats while maintaining the model utility and ensuring data privacy, contributing to the overall system integrity.  
\subsubsection{Defenses against Malicious Servers}
Malicious servers represent a dual threat to client privacy and overall system robustness~\cite{hao2023robust, han2024privacy}. They may engage in both passive and active attacks, often analyzing individual client updates or isolating shared models to compromise the underlying security model~\cite{Jeter2023OASIS, 10571602}. 
In addition to typical snooping, servers can utilize advanced multitasking strategies to create adversarial networks intended for client identification, which directly infringe user privacy~\cite{10184496, mothukuri2021survey}.
The server's ability to intercept personal data and deduce confidential information by closely analyzing model predictions or gradients~\cite{guo2023fast} further exacerbates this risk.
Furthermore, by introducing malevolent parameters, crafted gradients, or corrupted updates to undermine accuracy~\cite{han2024privacy}, a compromised server can intentionally impair the global model's performance. Such manipulation extends beyond performance measures; a malevolent server can add systemic bias and compromise the fairness of the global model by actively altering model weights or parameters~\cite{tang2024flexible}. Studies that rely on decentralized and traceable blockchain~\cite{zhou2020pirate, ma2022federated} have shown remarkable performance in handling these threats. To prevent malicious servers from manipulating global model parameters to infer private data,~\cite{Jeter2023OASIS} developed an \textit{image augmentation}- based defense that preserves model performance while showing strong resilience against such attacks.  \text{Enhanced ciphertext strategies} also provide defense against malicious server and client threats~\cite{10571602}.  
\subsubsection{Adversarial-resistant Aggregation} 
Several techniques have been proposed to combine model updates from multiple participants while mitigating the impact of malicious or deceptive inputs.
We dedicated the entire Section~\ref{sec:aggr} to discussing the robust aggregation strategies. 
\subsubsection{Advanced Cryptographic Strategies}
In addition to ensuring the integrity and confidentiality of the aggregation process, the integration of cryptographic strategies has also emerged as a valuable tool to enhance both security and robustness against sophisticated threats such as poisoning attacks~\cite{xia2023poisoning, sotthiwat2021partially, zhu2021distributed, ma2024vpfl}. 
The lack of transparency in the FL training process, due to its localized nature, makes it difficult for third parties, such as model users or auditors, to verify the model's integrity and correctness. 
 \textit{Zero-knowledge proofs} (ZKPs) address this issue by allowing public verification of the learning process without disclosing confidential information about the model or the data~\cite{ma2024vpfl}.
Moreover, \textit{Post-Quantum Cryptography (PQC)}~\cite{javeed2024quantum}, \textit{quantum cryptographic approaches} are becoming increasingly popular.
In contrast to conventional RSA and ECDSA, which are susceptible to Shor's algorithm, PQC is based on a mathematical foundation that can resist quantum-scale attacks~\cite{javeed2024quantum, chehimi2022quantum}. To resist emerging quantum threats, \cite{gurung2023performance} integrated PQC within \textit{blockchain-based FL} by using a combination of stateful and stateless hash-based signature mechanisms. Similarly, \textit{HE} has been used as a defense mechanism against poisoning threats. For instance, the \textit{two-trapdoor HE} mechanism devised in~\cite{ma2022shieldfl} successfully identifies and eliminates malicious updates by estimating distances between encrypted gradients using secure cosine similarity. This ensures robustness while preserving privacy. This method is particularly effective in heterogeneous scenarios and can withstand up to 50\% adversarial users. Even though popular schemes such as El Gamal, RSA, and Paillier exhibit homomorphic characteristics~\cite{zhu2021distributed}, applying them to FL faces a number of trade-offs, such as increased memory overhead, latency, and polynomial approximations for non-linear models~\cite{moriai2019privacy}. \textit{polymorphic encryption} (PE) has also been experimented in this context to resist inference and poisoning attacks~\cite{moshawrab2023polyflag_svm}. 
\subsection{Key Insights on Defensive Strategies}
We present Table~\ref{tab:defenses-summary} to provide a quick summary of robustness through defensive mechanisms, and Table~\ref{tab:def-insights} to deliver the key insights.
The analysis demonstrates that upholding robustness demands layered defenses covering data, updates, adversarial aggregation, global model verification, server behavior analysis, and cryptographic safeguards. AEs, DP, KD, and client-side preprocessing work together to reduce \emph{inference risk and poisoning} while maintaining accuracy in the face of training data manipulation. Although they often come at the expense of additional computation, validation data, or protocol complexity, \emph{robust aggregation, similarity/trust mechanisms, truth discovery, sparsification, autoencoders, watermarking, and blockchain} provide additional protection against malicious model updates. 
Vital sanity-checks should be performed to \emph{ensure global integrity} (such as Eiffel, decentralized aggregation, trust-validation, and personalized encoders) to preserve or restore a clean global model in the face of severe attacks. By reducing reconstruction, client identification, and biased aggregation, defenses against \emph{malicious servers}, such as image augmentation, enhanced ciphertext, and blockchain-based auditing, specifically target the most powerful adversary. Moreover, strong confidentiality and verifiability are provided by \emph{sophisticated cryptographic techniques}, but these techniques can also incur additional overhead.
Thus, practical robust FL deployments should consider \emph{carefully selected hybrids} rather than relying solely on a single defense mechanism.
\begin{table*}[!htb]
\centering
\vspace{0.1in}
\caption{Key insights on robustness via defensive strategies}
\label{tab:def-insights}
\vspace{-0.1in}
\renewcommand*{\arraystretch}{1}
\resizebox{1\textwidth}{!}{
\begin{tabular}{|m{2.2cm}|m{5cm}|m{5cm}|m{5cm}|m{6cm}|m{6cm}|}
\hline
\rowcolor[HTML]{8c8c8c}
\textbf{Insight} & \textbf{Core idea} & \textbf{Defended threats} & \textbf{Typical design pattern} & \textbf{Advantages} & \textbf{Drawbacks} \\
\hline
Conventional defenses fail in FL
&
Classical centralized defenses that assume direct access to training data and full control over the pipeline conflict with FL’s distributed approach.
&
Data poisoning in raw data, data-level outliers, and direct anomaly detection at the sample level.
&
Assume the server sees the full dataset; reweight or discard suspicious samples using robust losses or feature statistics.
&
Well-understood in centralized ML; strong guarantees only when data is pooled.
&
Incompatible with FL’s update-only interface, violates privacy constraints, and cannot inspect raw local data or intermediate features.
\\
\hline
Multi-attack coverage
&
Defenses that simultaneously address multiple threats appear more appealing than mechanisms tailored to a single attack type, motivating taxonomies organized by phase or location rather than by attack.
&
Label-flipping, gradient poisoning, backdoors, inference or reconstruction attacks.
&
A combination of robust aggregation with DP noise, KD, autoencoder-based update scoring, or sparsification.
&
One mechanism spans multiple stages (local, aggregation, global) and minimizes attack-specific tuning, making it convenient in mixed-threat environments.
&
Difficult to ensure the best protection for any single attack, covert and adaptive adversaries can still circumvent generic rules, DP noise, and robustness interaction is non-trivial.
\\
\hline
Layered client-server protection 
&
Combined protection across local training, aggregation, and system layer for increased effectiveness than any isolated component.
&
Local poisoning; gradient leakage; compromised servers; integrity of global model and logs.
&
Client (preprocessing, DP, VAE, KD, etc.), server (similarity or trust scoring, robust aggregation), system (blockchain, auditing, watermarking).
&
Suppress attacks early at the client, preserve long-term traceability and accountability of updates, and model lineage.
&
Increased overall complexity, difficulty in coordinating thresholds and hyperparameters across layers, and overhead on resource-constrained devices.
\\
\hline
Cryptography boosts robustness
&
Strong cryptography enables confidentiality and verifiability, but is resource-intensive.
&
Inference attacks, model or gradient inversion, leakage, poisoning, and replay attacks.
&
Secure aggregation, ZKP,  PE, and  PQC techniques
&
Securely fetch individual updates from the server and peers, allow validation during aggregation and training, tolerate stronger threat models (e.g., curious or semi-malicious servers).
&
Encrypted computation is slow (high runtime),  nonlinear networks require polynomial approximations, and protocols are complex and involve key management issues.
\\
\hline
Fragile defensive assumptions
&
Simple assumptions that may break under sophisticated or more realistic conditions.
&
High-Byzantine poisoning, adaptive and covert adversaries that target defense logic, highly non-IID clients.
&
Assume a maximum fraction of suspicious clients, need good validation data or honest clients, threshold tuning for specific non-IID levels.
&
Offer provable bounds or strong empirical robustness under an assumed context, easy to analyze and implement for targeted scenarios.
&
Bound on malicious fraction; need auxiliary data or TEE; degrade under extreme non-IID.
\\
\hline
Robustness through personalization
&
Client or cluster-specific models to reduce the impact of poisoned or biased  updates.
&
Client-targeted poisoning; non-IID amplified attacks; fairness/bias issues.
&
Clustered aggregation, personalized layers,  trust-weighted personalization, KD-based refinement, representation or prototype-level adaptation.
&
Reduces cross-client poisoning; isolates malicious clusters, better fits local data, improves accuracy and robustness on non-IID data.
&
More models to manage, difficulty in global guarantees, and misassignment of clusters strengthen attacks.
\\
\hline
Ownership and auditability
&
Verify training integrity and prevent theft, copying, or claiming ownership of the model.
&
Model theft, unauthorized model re-use, disputes about training data or protocol.
&
Watermarking, blockchain (append-only logs for updates), signed update histories, ZKPs for verifiable training.
&
Enables model origin verification and tamper checks, easier detection of adversarial modifications, and supports audit requirements.
&
May not prevent attacks in real time; a delicate trade-off between watermark robustness and accuracy; a blockchain or logging approach incurs an operational burden.
\\
\hline
Future-guarantees with PQC and hybrid designs
&
Against future quantum attackers while staying practical via hybridization tactics.
&
Update confidentiality, key compromise, signature forgery, and long-term model poisoning.
&
PQC-based signatures,  PQC + blockchain, PQC-secured aggregation, lightweight crypto integration with statistical defenses (DP, robust aggregation, sparsification).
&
Security against emerging cryptanalytic capabilities, verifiable logs, and secure channels, integration with existing FL pipelines.
&
PQC tooling is in nascent stages, larger keys and signatures, hybrid design and deployment remain complex.
\\
\hline
\end{tabular}
}
\vspace{-0.1in}
\end{table*}
\section{Experimental Evaluation Trends} \label{sec:cri}
This section analyzes current trends in the experimental evaluation of robust FL systems, focusing on data partitioning approaches and the datasets commonly used for robustness testing. 
The analysis highlights the strengths and weaknesses of current methodologies and provides insights into potential areas that demand more robust evaluations.
\subsection{Data Partitioning Approaches}
    Evaluating FL algorithms often involves simulating non-IID data distributions across clients, even when the main focus is on handling heterogeneity rather than treating it as an additional source of noise.
    Meanwhile, for IID data partitioning, the data is randomly and evenly distributed across clients, ensuring each client has a similar distribution.
    The most common approach to simulate non-IID conditions is the \textbf{Class-based approach}. Each client typically receives data corresponding to a limited number of classes, with uneven sample counts, reflecting a highly skewed distribution. For instance, in~\cite{jeong2018communication}, 2000 samples are selected randomly and divided into 10 subsets based on their true labels. Each client is then assigned a fixed number of target labels. On the other hand, in~\cite{briggs2020federated}, clients are given samples with only two labels, with each client receiving 600 instances. Another common strategy is the Dirichlet distribution-based approach, where data is allocated to clients based on a Dirichlet distribution, allowing for varying degrees of non-IIDness.
The \textbf{Category-based approach}, instead, performs distribution followed by local classification. The data slices are distributed across clients, and classification is performed on the client side. For instance, in~\cite{zhao2018federated}, the data is sorted by category into 20 partitions, and each client randomly receives two partitions from the 2 categories.
Although these approaches are commonly considered standard, the majority of studies typically consider a single non-IID setting and rely on relatively simple varieties of label skew on vision benchmarks, overlooking stronger versions of volume, label, and feature skews.  For instance, the same defense can behave very differently depending on the partition choice. Assessing robustness under dynamic or time-varying trends that more accurately reflect real deployments, combining multiple skew types, and systematically varying distribution skew (e.g., varying Dirichlet concentration) are potential areas that warrant further attention.
\begin{table*}[!htb]
\renewcommand{\arraystretch}{1}
\centering
\vspace{0.1in}
\caption{Summary of commonly used datasets in FL research.}
\vspace{-0.1in}
\Description{Common datasets used by researchers in FL}
\resizebox{1\textwidth}{!}{
\begin{tabular}{|m{2.5cm}|m{4cm}|m{8cm}|m{0.6cm}|m{1.2cm}|m{8cm}|} \hline
      \rowcolor[HTML]{8c8c8c}  \textbf{Dataset} & \textbf{Type} & \textbf{Description} & \textbf{IID} & \textbf{Non-IID} & \textbf{Applied Scenario} 
\\ \hline
CIFAR-10
& Image Classification & 60,000 32x32 color images across 10 classes & $\checkmark$ & $\checkmark$ & Object recognition, autonomous systems. \\ \hline
 CIFAR-100
 & Image Classification & 60,000 32×32 color images across 100 classes & $\times$ & $\checkmark$ & Fine-grained classification, FL model evaluation purposes. \\ \hline
 MNIST
 & Handwritten Digits & 70,000 28×28 grayscale images (digits 0-9) & $\checkmark$ & $\checkmark$ & Handwritten digit recognition, basic FL benchmarking. \\ \hline
Extended MNIST
& Handwritten Characters & Includes handwritten letters and digits &  $\times$ & $\checkmark$ & Real-world non-IID settings for character recognition tasks. \\ \hline
Fashion-MNIST
& Fashion Image Classification & 70,000 28×28 grayscale images across 10 fashion categories & $\checkmark$ & $\checkmark$ & Retail analytics, product categorization. \\ \hline
Shakespeare
& Text (NLP) & Complete works of Shakespeare divided by characters &  $\times$ & $\checkmark$ & Personalized language modeling, and text generation tasks. \\ \hline
Google's Gboard
& Keystroke Data (NLP) & Text input data from Google's Gboard keyboard &  $\times$ & $\checkmark$ & Next-word prediction, language modeling. \\ \hline
Application-specific & Application-related Data (e.g., Medical, Speech, IoT) & Domain-specific datasets for various real-world applications &  $\times$ & $\checkmark$ & Healthcare, IoT, Speech recognition.  \\ \hline
Synthetic & Simulated Data & Artificially generated datasets for algorithm testing & $\checkmark$ & $\checkmark$ & Experimenting FL model validation under controlled conditions and contexts. \\ \hline
Others & Miscellaneous & Other miscellaneous datasets& $\checkmark$ & $\checkmark$ & FL research on custom datasets. \\
\hline
\end{tabular}}
\label{tab:dataset}
\vspace{-0.2in}
\end{table*}
\subsection{Common Datasets in FL Research}
\begin{figure}[!htb]
\vspace{-0.2in}
    \centering
       \begin{subfigure}[b]{0.49\linewidth}  
        \centering
        \includegraphics[width=\linewidth]{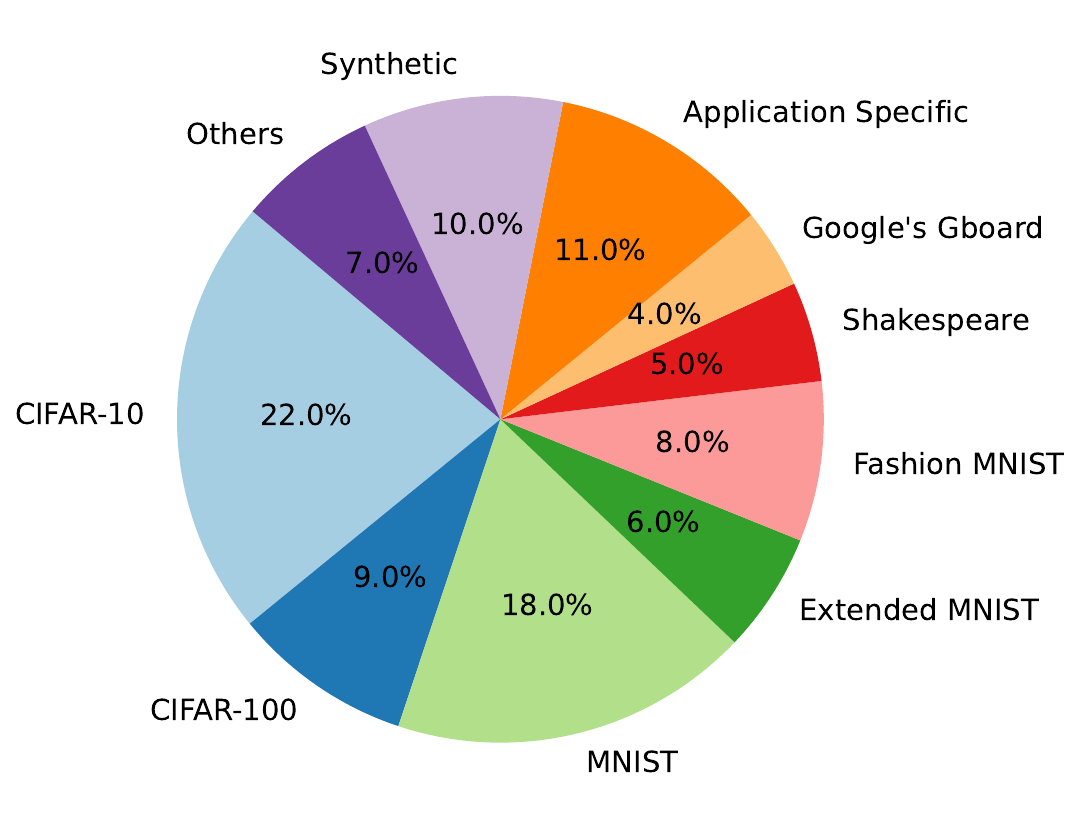}
        \caption{Commonly used datasets in FL research.}
        \Description{FL datasets used by researchers.}
        \label{fig:dataset}
    \end{subfigure}
    \hfill
    \begin{subfigure}[b]{0.49\linewidth} 
        \centering
        \includegraphics[width=\linewidth]{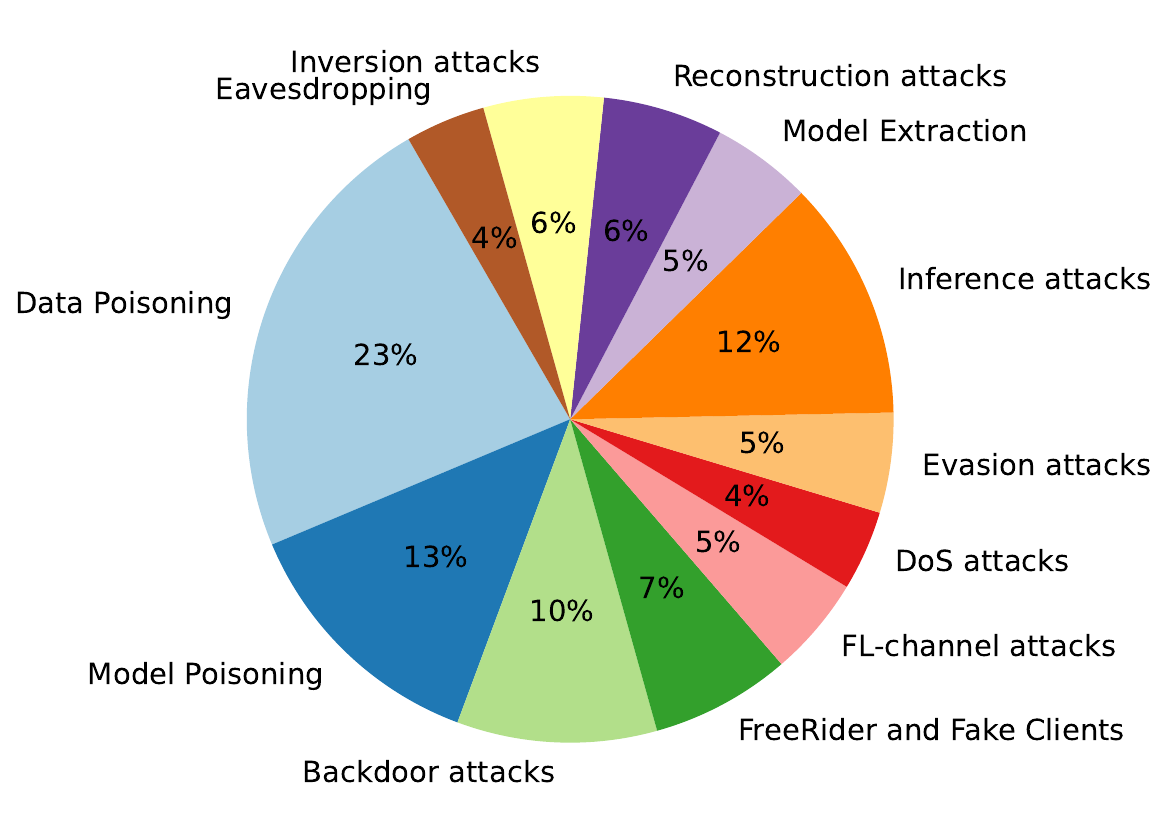}
        \caption{Research focus across FL attack categories.}
        \Description{Fl research focus in terms of attacks}
        \label{fig:resfocus}
    \end{subfigure}    
    \caption{Current Research Trends in Robust FL.}
    \vspace{-0.1in}
    \label{fig:publications}
\end{figure}

Unless an application-specific dataset is used, researchers in FL typically rely on a set of commonly used datasets to evaluate the performance and robustness of their algorithms. 
The most commonly used datasets are summarized in Table~\ref{tab:dataset}, along with their type, description, suitability for data distribution, and real-world application scenarios. 
Fig.~\ref{fig:dataset} represents the distribution of commonly used datasets in FL research.  
CIFAR-10 and MNIST remain foundational, demonstrating their continued relevance in collaborative model training scenarios. Recently, the number of works using Fashion-MNIST for image classification has increased. 
\subsection{Research Trends in Robust FL}
According to the threat modeling literature, researchers have mostly focused on a few well-known attack types, including backdoor attacks, model poisoning, and data poisoning. Given their direct impact on the model's integrity, they have been extensively explored by researchers. Other key adversarial threats include inference attacks, free-riding and malicious behavior, FL-channel attacks, evasion strategies, and network-level threats such as eavesdropping and DoS attacks. Fig.~\ref{fig:resfocus} portrays the major research focus across various FL attack categories.
The current literature is dominated by works that either analyze the attack surface or design concrete defense mechanisms. 
While defense mechanism focuses on countering diverse threats with customized protective strategies at the client or server side, attack surface analysis primarily focuses on methodically characterizing threats to robustness in FL. In comparison, the researchers paid less attention to aggregation reliability and heterogeneity. This indicates that FL robustness is still largely influenced by sophisticated adversarial circumstances, necessitating increased focus. 
\section{Discussion and Future Directions}\label{sec:future}
This section discusses key insights into emerging trends to robustness in FL, followed by the major application scenarios of robust FL, challenges faced, and future research directions.
\subsection{Emerging Robustness Trends in FL}
FL is still evolving, with a number of new trends reshaping its overall robustness in dynamic environments, deployment efficiency, and design concept. The demand to improve scalability, communication efficacy,  privacy, and model robustness in heterogeneous and dynamic environments is driving innovations. A summary analysis of emerging research trends in robust FL is presented in Table~\ref{tab:trends}. The goal of new methods, such as communication scheduling,  model pruning, and asynchronous updates, is to reduce communication costs, especially in devices with constrained resources~\cite{jiang2023complement, 10571602}. In order to improve data privacy and guard against potential quantum threats, sophisticated cryptographic solutions like HE and PQC are being incorporated as well~\cite{javeed2024quantum, gurung2023performance}. 
Hierarchical and clustered architectures and lightweight learners further support scalability, and personalization techniques guarantee that client-based requirements are guaranteed without undermining global performance \cite{jiang2020improving, li2022federated, caldarola2021cluster, briggs2020federated, collins2021exploiting}. 
Recently, investigations into Quantum ML and quantum-influenced optimization have opened innovative avenues to advance model training and resilience \cite{chehimi2022quantum, zhang2022federated, yang2022post}. Altogether, these rising trends indicate a positive shift toward more robust, flexible, and future-ready FL frameworks. 
\begin{table}[!htb]
\centering
\small
\renewcommand{\arraystretch}{1}
\caption{Emerging research trends toward Robust FL.}
\vspace{-1em}
\resizebox{1\linewidth}{!}
{\begin{tabular}{|m{7cm}|m{10cm}|}
\hline
\rowcolor[HTML]{8c8c8c} \textbf{Focus Area} & \textbf{Key Robust Strategies} \\
\hline
Strengthening Privacy and Resilience & 
- Differential Privacy with noise injection \newline
- Homomorphic and polymorphic encryption \newline
- Post-Quantum Cryptography (PQC) \newline
- Blockchain and SMC for decentralized trust \\
\hline
Designing Scalable Federated Architectures & 
- Lightweight learners (e.g., ELM) for edge nodes \newline
- Hierarchical, cluster-based FL frameworks \newline
- Decentralized aggregation to mitigate central bottlenecks \newline
- Quantum ML and Hybrid Classical-Quantum Systems \newline
- Asynchronous FL to control latency and device churn. \\
\hline
Advancing Personalization and Adaptivity & 
- Personalized models using meta-learning and local fine-tuning. \newline
- Use of adaptive optimizers for faster convergence \newline
- Multi-task learning for promoting personalized models. \newline
- Client-specific loss functions or model layers to support heterogeneity  \newline
- Semi-supervised and multimodal FL approaches \\
\hline
Enhancing Robustness and Model Integrity & 
- Robust aggregators to deal with malevolent or corrupted updates \newline
- Reputation-driven solutions for client selection  \newline
- Quantum-driven optimization \newline
- Fault tolerance against participant dropouts and stragglers \newline
- Dynamic selection of clients based on device availability and reliability. \\
\hline
Robust Communication Efficiency & 
- Model pruning, gradient compression to reduce overhead \newline
- Sparsity and quantization for efficient aggregation  \newline
- Asynchronous and layer-wise updates \newline
- Communication scheduling based on device capacity \\
\hline
\end{tabular}}
\vspace{-0.3in}
\label{tab:trends}
\end{table}
\subsection{Major Application Scenarios for Robust FL }
This section explores major real-world FL application scenarios across different sectors, where robustness remain critical yet underexplored challenge.
\subsubsection{Intelligent Healthcare}
In healthcare, safeguarding patient privacy is paramount, but this should be achieved without sacrificing the robustness and dependability of clinical models. FL enables multiple hospitals and clinics to collaboratively train models without sharing patient data externally. To guarantee reliable decision support in safety-critical workflows, robust FL must also withstand noisy labels, diverse patient populations, and local updates poisoning, among others. To improve robustness against data manipulation and single points of failure, decentralized aggregation and coordination components (e.g., committee-based or blockchain-assisted aggregation) are used.  The healthcare-FL model in~\cite{ye2023heterogeneous} predicts oxygen requirements using vital signs, laboratory reports, and chest X-rays to improve clinical outcomes
A diagnostic model using FL in~\cite{ma2022assisted} supports cancer patients by letting physicians to design personalized nutrition and treatment plans, support longer life expectancy, and offer guidance in rehabilitation. 
\subsubsection{Recommender Systems}
 Robust federated recommendation offloads computation to user devices while shielding the learning and aggregation pipeline against such risks. Recent advances include anomaly-resilient aggregation, client trust scoring, and privacy-preserving protections to prevent adversaries from extracting sensitive preferences or influencing recommendations. Different approaches include: (i) Robust collaborative filtering: Utilizing user-item interaction data while preserving privacy by handling user-item factors under adversarial ratings and sparse or skewed interaction graphs, often using outlier-resistant aggregators, (ii) Deep learning-based robust FL: Employing neural networks trained across distributed clients, while using robust aggregation, gradient masking, or client selection approaches to mitigate adversarial risks, (iii) Meta-learning based robust FL: Personalizing models for individual users through experience-based adaptations to mitigate malicious or low-quality clients.
A distributed matrix decomposition approach~\cite{du2021federated} helps learning global latent factors via gradient sharing, and integrating matrix factorization with quantization helps minimize communication overhead.
FL-driven recommender systems support a variety of sectors, including e-commerce and social media, in providing customized services while adhering to data regulations and privacy issues. However, overlooking robustness could allow malevolent alterations of recommendations, leading to reduced user confidence, user discontent, and monetary losses~\cite{nguyen2026handling}.
\subsubsection{Finance and Banking}
Collaboration between financial institutions to enhance services such as fraud detection, credit risk analysis,  and financial recommendations is made possible by robust FL. Confidentiality challenges and competitive interests, however, limit explicit data exchange between institutions. The initial results are encouraging and reveal FL's potential to improve data-driven choices for decision-making across financial networks, even though the majority of FL-driven financial applications are currently in their infancy. Robustness preserves confidentiality and integrity of the model and is guaranteed through threat-resilient optimization, secure aggregation,  and cryptography-oriented defenses.
Notable robust applications include: (i) WeBank's credit risk assessment~\cite{li2023research} uses FL for small business lending and personal credit scoring, supporting credit risk management, mitigating single-institution bias, and limiting the impact of corrupted or low-quality clients, (ii) FL for  Internet banking in~\cite{luo2023application} integrated DP, SMC, and collaborative modeling for improving credit approval rates and loan performance while ensuring data confidentiality.
\subsubsection{Smart Cities}
Smart cities require integrating and analyzing data generated by diverse entities, including governments, private enterprises, and individual users. Robust FL enables decentralized training of AI models across urban stakeholders while mitigating malicious clients, corrupted sensor devices, and unreliable communication, thereby supporting safety-critical decision-making. This supports various urban applications, such as traffic prediction, pollution monitoring, and infrastructure management~\cite{que2025scalable, li2022federated}. For instance, the urban traffic prediction framework in~\cite{li2022federated} employs a global spatial-temporal pattern graph under a data federation, ensuring data privacy using personalized FL methods based on meta-learning while mitigating the impact of heterogeneous and noisy data sources. Historical traffic statistics is employed in ~\cite{yuan2022fedstn} for congestion prediction, while ensuring privacy. 
\subsubsection{IoT and Edge Computing}
FL and edge computing combination enables distributed model training directly on edge devices, such as smartphones, smart meters, surveillance cameras, and industrial IoT nodes, without transferring raw data to a central cloud. This preserves data privacy and strengthens security while reducing communication overhead and latency~\cite{que2025scalable, zhang2022fedada, yuan2022fedstn}. Handling malevolent updates, heterogeneity of devices, and unstable connectivity is vital for applications such as industrial supervision and surveillance, where compromised models can have serious consequences.
Numerous studies have looked at different aspects of this integration. For instance,
local training on mobile devices is enabled by an edge FL framework in~\cite{ye2020edgefed}, which allows for robust server-side inspection of anomalous updates and periodically aggregates model updates to enhance learning efficiency and reduce overhead. A fair aggregation strategy can ensure balanced accuracy among clients and reduce the dominance of low-resource or unreliable clients. An FL model with adaptive learning rates~\cite{jiang2020customized} addresses the various accuracy needs of heterogeneous edge devices. By down-weighting distrustful contributions, an asynchronous FL model in~\cite{wang2021efficient} improves communication efficiency through selective model updates based on device capacity and data quality. 
\subsection{Challenges and Future Research Directions}
While current robust mechanisms have addressed fundamental threats and challenges facing FL, the evolution of adversarial tactics necessitates more sophisticated, flexible, scalable, and domain-aware solutions. The next frontier for robustness in FL is discussed along the following directions:
\noindent $\bullet$ \textbf{Causality-driven attack attribution:}  Current detection measures typically struggle to distinguish between malicious model poisoning and benign statistical heterogeneity. 
    Future studies should adopt the use of causal inference to distinguish spurious correlations from adversaries, allowing for more precise detection and reliable source attribution.
\noindent $\bullet$ \textbf{Context-aware and flexible defenses:} The adoption of fixed or static parameters for defenses (e.g., preset clipping thresholds) is often unreliable. Robust FL demands adaptive frameworks that incorporate temporal, environmental, and contextual factors, tweaking the level of defenses in real-time according to the shifting threat landscape, client behavior trends, and application-based priorities.
\noindent $\bullet$ \textbf{Self-healing and resilient frameworks:} Apart from mere detection strategies, FL frameworks must have self-healing capabilities. Drawing motivation from biological immunity, subsequent research should focus on innovative solutions that enable the models to behave automatically according to any recognized degradation in performance, separate infected sub-modules, and activate automated recovery appropriately without needing a system reset.
\noindent $\bullet$ \textbf{Neuro-symbolic FL for verifiable robustness:}  Adversarial examples often affect purely neural approaches. To address this, integrating symbolic-AI with neural models that can provide verifiable reasoning and logical constraints is essential. It is particularly critical for robustness in high-stakes environments such as healthcare or autonomous navigation, where model decisions must adhere to strict safety logic.
\noindent $\bullet$ \textbf{Cross-domain and modality-specific defenses:}  Image classification is the subject of a large portion of the existing literature. However, there are significant differences in the robustness requirements for different forms of data. Future research must devise robustness primitives specifically designed to address the distinct structural weaknesses of non-vision sectors as well.
\noindent $\bullet$ \textbf{Quantum-enhanced and quantum-resilient FL:} (i) \textit{Quantum FL (QFL):} Effective estimation and addressing gradient vanishing in deep quantum circuits is vital. Future studies should explore quantum-based compression to reduce communication overhead and personalized QFL for resource-limited edge devices, which adversaries commonly exploit.
        (ii) \textit{Multimodal QFL:} Creating QFL systems that can accommodate the integration of different kinds of data will broaden their use in major applications, such as smart healthcare and finance.
        (iii) \textit{Quantum-Resilient security:} Many of the standard encryption solutions will turn obsolete as quantum computing advances. Transition of FL towards PQC is therefore crucial to ensure long-term data protection and integrity against adversaries with quantum capabilities.
We presented a multifaceted review and analysis of robust FL from diverse perspectives. Given the increasing sophistication of adversarial threats and the inherent challenges presented by the heterogeneous and decentralized characteristics of FL, the analysis has emphasized the critical relevance of upholding robustness in FL systems.
\vspace{-0.1in}
\bibliographystyle{ACM-Reference-Format}
\bibliography{survey}
\end{document}